\pdfoutput=1

\documentclass[11pt]{article}

\usepackage[]{ACL2023}

\usepackage{times}
\usepackage{latexsym}

\usepackage[T1]{fontenc}

\usepackage[utf8]{inputenc}

\usepackage{microtype}
\usepackage{booktabs}
\usepackage{multirow}
\usepackage{multicol}
\usepackage{lipsum}
\usepackage{graphicx} 
\usepackage{amssymb}
\usepackage[table]{colortbl}
\usepackage{amsmath}
\definecolor{darkgreen}{HTML}{3C8031}
\usepackage{stfloats}

\newcommand{\adhoc}{\textsc{MiniJoint}}
\newcommand{\posthoc}{\textsc{MiniPost}}
\newcommand{\baselinelarge}{\textsc{BL\_Base}}
\newcommand{\baselinesmall}{\textsc{BL\_Small}}

\usepackage{xcolor,colortbl}
\usepackage{subcaption}
\definecolor{Gray}{gray}{0.90}
\newcolumntype{a}{>{\columncolor{Gray}}c}

\title{Mini-Model Adaptation: Efficiently Extending Pretrained Models \\ to New Languages via Aligned Shallow Training}

\newcommand{\affilsup}[1]{\rlap{\textsuperscript{\normalfont#1}}}

\newcommand*\samethanks[1][\value{footnote}]{\footnotemark[#1]}

\author{
    Kelly Marchisio\affilsup{1,2}\ \ \ \ \thanks{\ \ Work done during an internship at Meta AI} \qquad
    Patrick Lewis\affilsup{2}\ \ \thanks{\ \ Work done at Meta AI} \qquad
    Yihong Chen\affilsup{3,4} \qquad
    Mikel Artetxe\affilsup{5} \ \samethanks \\
    $^1$Johns Hopkins University \qquad
    $^2$Cohere AI \qquad
    $^3$Meta AI \\
    $^4$University College London \qquad
    $^5$Reka AI \\
    \texttt{kmarc@jhu.edu, mikel@reka.ai}
}

\begin{document}
\maketitle
\begin{abstract}
Prior work shows that it is possible to expand pretrained Masked Language Models (MLMs) to new languages by learning a new set of embeddings, while keeping the transformer body frozen. Despite learning a small subset of parameters, this approach is not compute-efficient, as training the new embeddings requires a full forward and backward pass over the entire model. 
We propose \textit{mini-model adaptation}, a compute-efficient alternative that builds a shallow \textit{mini-model} from a fraction of a large model's parameters. New language-specific embeddings can then be efficiently trained over the mini-model and plugged into the aligned large model for rapid cross-lingual transfer. 
We explore two approaches to learn mini-models: \adhoc{}, which jointly pretrains the primary model and the mini-model using a single transformer with a secondary MLM head at a middle layer; and \posthoc{}, where we start from a regular pretrained model, build a mini-model by extracting and freezing a few layers, and learn a small number of parameters on top. Experiments on XNLI, MLQA and PAWS-X show that mini-model adaptation matches the performance of the standard approach using 2.3x less compute on average.
\end{abstract}

\section{Introduction}
\label{sec:introduction}

\begin{figure}[t]
\centering
\includegraphics[width=1\linewidth,trim={1cm 0.5cm 1cm 1cm},clip]{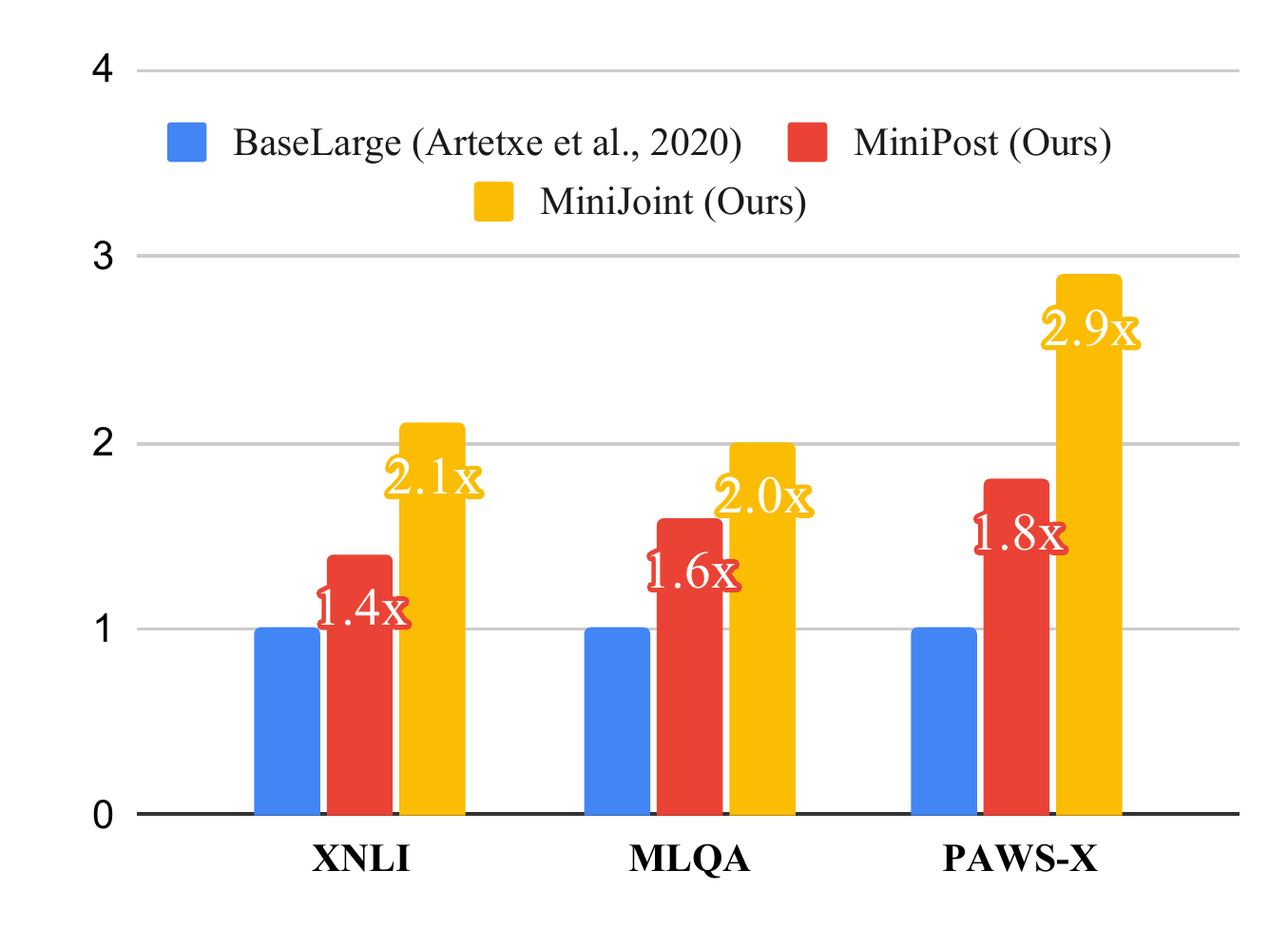}
\caption{\textbf{Average speedup of mini-model adaptation over \citet{artetxe-etal-2020-cross}}. A speedup of $x$ means that our approach needs $x$ times less compute to achieve the same performance. See \S\ref{sec:near-max} for more details.}
\label{fig:speedup}
\end{figure}

Recent work on multilingual NLP has focused on pretraining (masked) language models on unlabeled corpora in multiple languages \citep{pires-etal-2019-multilingual,conneau-etal-2020-unsupervised,xue-etal-2021-mt5}. The resulting models can then be finetuned using labeled downstream data in a single language (typically English), and zero-shot transferred to the rest of the languages.
While effective, existing models rarely cover more than a few dozen languages, and pretraining new models from scratch to support additional languages can be prohibitively expensive.

Motivated by this, a recent line of work has explored pretraining an initial model in a few languages, and expanding it to new languages post-hoc in a continual learning fashion \citep{mhamdi2022crosslingual}. More concretely, \citet{artetxe-etal-2020-cross} showed that it is possible to expand an English masked language model (MLM) to new languages by freezing the transformer body and learning a new embedding layer using the original MLM objective. 
Recent work has reported improved results by using a better initialization scheme \citep{pfeiffer-etal-2021-unks}, or learning additional language-specific parameters through adapters \citep{pfeiffer-etal-2022-lifting}. All these approaches are parameter-efficient, as they only learn a small number of parameters for each language, while the rest remain frozen. However, learning such parameters is not compute-efficient, as it requires a full forward and backward pass over the entire model, including the frozen transformer body.

\begin{figure*}[t]
\centering
\includegraphics[height=0.41\textheight]{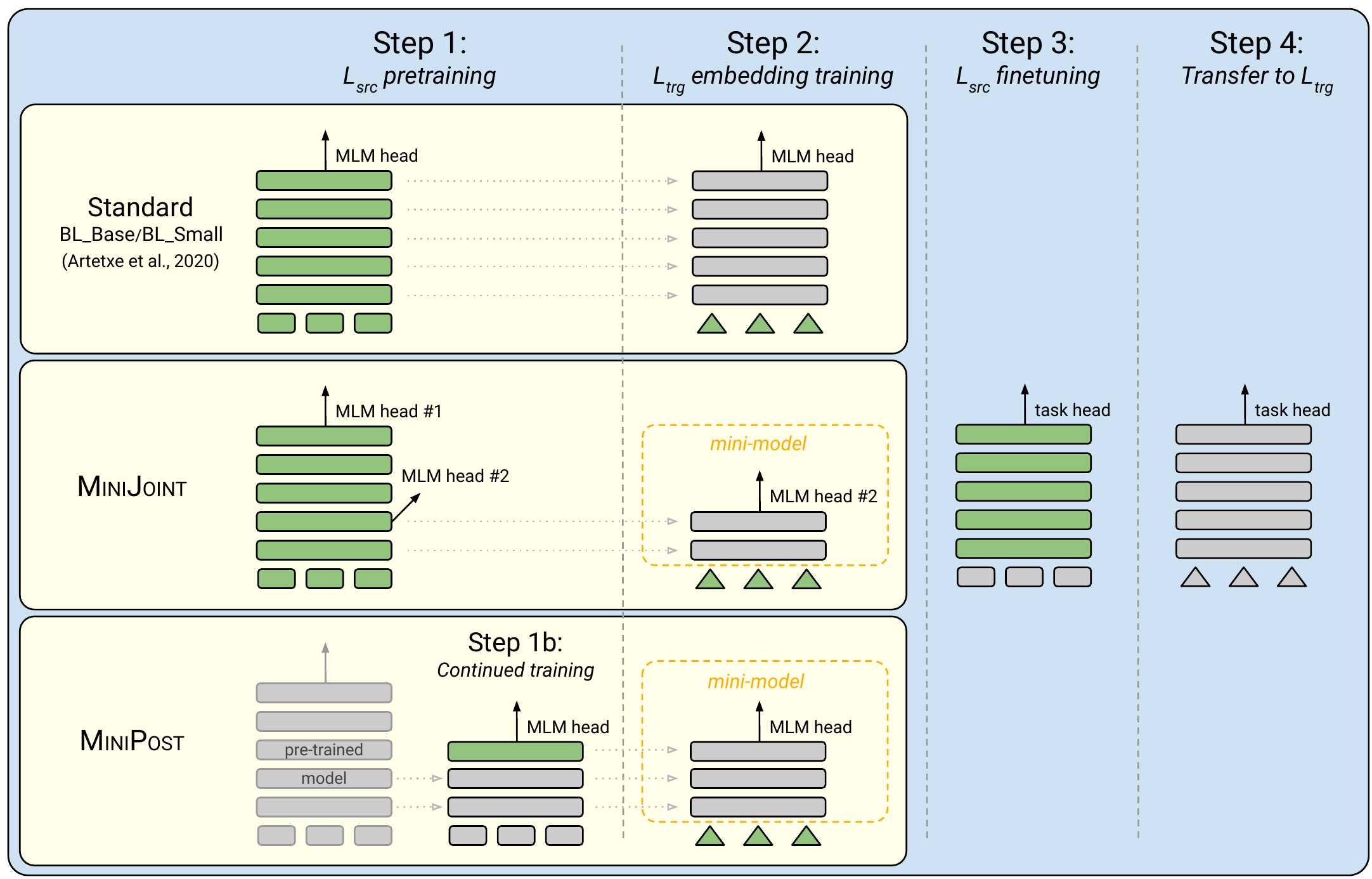}
\caption{
\textbf{Standard and mini-model adaptation.}
Trainable parameters are \textit{\textcolor{darkgreen}{green}}, frozen parameters are 
\textit{\textcolor{gray}{gray}}. $L_\text{src}$ embeddings are small rectangles, $L_\text{trg}$ embeddings are triangles.
All approaches use a four-step process for cross-lingual transfer: (1) pretrain an MLM in $L_\text{src}$, (2) learn a new embedding layer in $L_\text{trg}$ via MLM with transformer body frozen, (3) finetune the model in $L_\text{src}$ with embeddings frozen, (4) zero-shot transfer to $L_\text{trg}$ by swapping the embeddings. \textbf{Standard adaptation (top)} uses the same transformer body for all steps, while our approach learns two aligned models in Step 1---the primary model and a shallower mini-model---and uses the mini-model to learn $L_\text{trg}$ embeddings efficiently in Step 2. We explore two approaches to learn mini-models: \textbf{\adhoc{} (center)} jointly pretrains the primary model and mini-model using a secondary MLM head attached at a middle layer; \textbf{\posthoc{} (bottom)} starts from an existing model and builds a mini-model in Step 1b by extracting/freezing a few layers and learning a small number of parameters on top.
}

\label{fig:methods}
\end{figure*}

We introduce \textit{mini-model adaptation}, a new approach to extend MLMs to new languages that is both parameter- and compute-efficient. \textit{Mini-models} are shallow models that are aligned with a larger parent model. Thanks to this, one can efficiently train a new embedding layer for a new language over the mini-model, and plug it directly into the parent for strong cross-lingual performance.

As shown in Figure \ref{fig:methods}, we explore two approaches to learn mini-models, depending on whether we start from an existing primary model and learn a mini-model posthoc (\posthoc{}), or we jointly learn the primary model and the mini-model from scratch (\adhoc{}). In \posthoc{}, we extract the bottom layers from the existing MLM, and learn a small number of parameters on top to make it a usable small MLM itself. In the \adhoc{} variant, we pretrain an MLM from scratch including a secondary head at a middle layer. Both heads are optimized jointly, creating a complete, well-aligned MLM contained within a larger MLM.

We evaluate our approach on natural language inference (XNLI), question answering (MLQA) and paraphrase identification (PAWS-X). As shown in Figure \ref{fig:speedup}, mini-model adaptation can match the performance of the standard method from \citet{artetxe-etal-2020-cross} using 1.6x and 2.3x less compute for \posthoc{} and \adhoc{}, respectively (averaged over tasks), and retains >98\% of performance when trained to completion.

All in all, our work shows that it is possible to adapt language models to new tasks (in this case, new languages) using smaller aligned models for training. While we focus on the problem of cross-lingual lifelong learning to validate this idea, we believe that this new paradigm opens exciting opportunities to make finetuning large language models more affordable.

\section{Proposed method}
\label{sec:method}

\subsection{Standard Adaptation} \label{subsec:baseline}

\citet{artetxe-etal-2020-cross} develop a four-step pipeline for cross-lingual transfer from a monolingual model, visualized in Figure \ref{fig:methods} (top).  First, one trains a monolingual MLM in the source language ($L_\text{src}$, usually English). Second, the transformer body is frozen, embeddings are re-initialized,\footnote{Following \citet{pfeiffer-etal-2021-unks}, we initialize the $L_\text{trg}$ embeddings with overlapping tokens from $L_\text{src}$ for all methods throughout. Non-overlapping tokens are randomly initialized using the normal distribution with  $\mu=0.0, \sigma=0.02$.} and the model is trained with MLM in the target language ($L_\text{trg}$). The trainable embeddings are tied with the output projection layer in the MLM head. Third, the $L_\text{src}$ embeddings are swapped back into the model and frozen, and the transformer body is finetuned on the downstream data in $L_\text{src}$.  Finally, the $L_\text{trg}$ embeddings are swapped back into the finetuned model for zero-shot transfer into $L_\text{trg}$.  We build two baselines based on this framework: a standard 12-layer (\baselinelarge{}), and a smaller 4-layer version (\baselinesmall{}).

\subsection{Mini-Model Adaptation} \label{subsec:method}

Our proposed approach follows a similar four-step training paradigm. However, we learn two aligned models in Step 1: the primary model and a shallow \textit{mini-model}. In Step 2, the $L_\text{trg}$ embeddings are learned over the mini-model, saving compute with respect to standard adaptation. Steps 3 and 4 are run as usual over the primary model, resulting in a full-size $L_\text{trg}$ model. For Step 1, we explore the following two alternatives depending on whether we start from an existing $L_\text{src}$ model, or we are training one from scratch:

\paragraph{\adhoc{}.} In this variant, we pretrain a dual-head 12-layer $L_\text{src}$ transformer from scratch, attaching a secondary head to an intermediary $N\text{th}$ layer (Figure \ref{fig:methods}, center). The model is trained to minimize the average MLM loss over the two heads. As such, the whole model receives gradient updates from the primary head, and the bottom layers also get updates from the secondary head. Having done that, we extract the bottom N layers and the secondary head to create the mini-model for Step 2. Unless otherwise indicated, we use $N\nolinebreak=\nolinebreak4$.

\paragraph{\posthoc{}.} Here, we start with a regular 12-layer MLM in $L_\text{src}$ (same as \baselinelarge{}), and build an aligned mini-model in Step 1b (Figure \ref{fig:methods}, bottom). To that end, we first copy the bottom $N$ layers into a new, shallower model, along with the embeddings and the MLM head. However, this does not work out of the box, as we must bridge the gap between the output of the $N$ bottom layers and the input of the MLM head, which goes through $12-N$ additional layers in the original model. To that end, we add 2 randomly-initialized layers between the $N$ bottom layers and the MLM head, and train them with the MLM objective in $L_\text{src}$ while keeping the rest of the parameters frozen. Because the new layers are unfrozen, they update to ``complete'' the MLM---bridging representations from the bottom layers' output to the MLM head's input, and resulting in a mini-model with $N+2$ layers that is fully functional and aligned with the primary model.

\section{Experimental Settings}
\label{sec:exp_settings}

\begin{table}[t]
\centering\footnotesize
\resizebox{\linewidth}{!}{
	\addtolength{\tabcolsep}{-3pt}
    \begin{tabular}{@{}lrlccc@{}}
    \toprule
       & GB    & Language Family            & \begin{tabular}[c]{@{}c@{}}Word \\ Order\end{tabular} & \begin{tabular}[c]{@{}c@{}}Syn.\\ Dist.\end{tabular} & \begin{tabular}[c]{@{}c@{}}Phylo.\\ Dist.\end{tabular} \\ \midrule
    ar & 28.0  & Afro-Asiatic: Semitic      & SVO/VSO & 0.57   & 1.00     \\
    bg & 58.0  & Indo-European: Slavic      & SVO     & 0.48   & 0.86     \\
    de & 67.0  & Indo-European: Germanic    & SVO/SOV & 0.42   & 0.43     \\
    el & 47.0  & Indo-European: Greek       & SVO/VSO & 0.52   & 0.83     \\
    en & 301.0 & Indo-European: Germanic    & SVO     & 0.00   & 0.00     \\
    es & 54.0  & Indo-European: Romance     & SVO     & 0.40   & 0.90     \\
    fr & 57.0  & Indo-European: Romance     & SVO     & 0.46   & 0.90     \\
    hi & 21.0  & Indo-European: Indic       & SOV     & 0.59   & 0.90     \\
    ru & 279.0 & Indo-European: Slavic      & SVO     & 0.49   & 0.90     \\
    sw & 1.7   & Niger-Congo: Bantu         & SVO     & 0.57   & 1.00     \\
    th & 72.0  & Tai-Kadai: Kam-Tai         & SVO     & 0.56   & 1.00     \\
    tr & 21.0  & Altaic: Turkic             & SOV     & 0.70   & 1.00     \\
    ur & 5.7   & Indo-European: Indic       & SOV     & 0.67   & 0.90     \\
    vi & 138.0 & Austro-Asiatic: Viet-Muong & SVO     & 0.57   & 1.00     \\
    zh & 47.0  & Sino-Tibetan: Chinese      & SVO     & 0.57   & 1.00     \\ 
    \bottomrule
    \end{tabular}
}
\caption{\textbf{Languages included in this study.}  GB: Size of CC-100 training data in gigabytes. Syn./Phylo. Dist.: syntactic and phylogenetic distance from English, respectively, according to \textit{lang2vec} \citep{littell2017uriel}.}
\label{tab:data}
\end{table}

\paragraph{Languages and Data.} Following common practice, we use English as the source language ($L_\text{src}$), and experiment with 14 other languages as the target ($L_\text{trg}$). We use CC-100 \citep{conneau-etal-2020-unsupervised} as our training corpus, which is a filtered version of CommonCrawl. We report the full list of languages along with their corpus size and linguistic details in Table \ref{tab:data}. Each language is preprocessed individually using SentencePiece \citep{kudo-richardson-2018-sentencepiece} with a vocabulary size of 50,000.

\paragraph{Models.}
We use the $\text{RoBERTa}_{\text{BASE}}$ \cite{liu2019roberta} architecture throughout from \textit{fairseq} \citep{ott-etal-2019-fairseq}.  Embeddings are tied.   
As said in \S\ref{sec:method}, we compare 4 systems: 2 variants of \citet{artetxe-etal-2020-cross} (\baselinelarge{} with 12 layers and \baselinesmall{} with 4 layers), and 2 variants of our proposed approach where we set $N\nolinebreak=\nolinebreak4$ (\adhoc{}, which jointly trains a 12-layer primary model and a 4-layer mini-model from scratch, and \posthoc{}, which starts from a regular 12-layer model and constructs a 6-layer mini-model post-hoc). \baselinelarge{} is a performance upper-bound, as it is the original 12-layer model that is used for adaptation. \baselinesmall{} is a lower-bound, demonstrating performance of the standard approach using an adaptation model of similar size as ours.

Models are trained for 125,000 steps with a global batch size of 2048, sequence length of 512, and learning rate of 7e-4 with 10,000 warmup updates and linear decay, both for the original pretraining (Step 1), and cross-lingual extension into each language (Step 2). As such, models see 131.1~billion training tokens per language. Step 1b in \posthoc{} uses the same training hyperparameters.

\paragraph{Evaluation.} We evaluate on 3 tasks: natural language inference in XNLI \citep{conneau-etal-2018-xnli}, question answering in MLQA \citep{lewis-etal-2020-mlqa}, and adversarial paraphrase identification in PAWS-X \cite{yang-etal-2019-paws}. We also report XQuAD \cite{artetxe-etal-2020-cross} results in \S\ref{sec:app-xquad}. In all cases, the model is finetuned using the corresponding training data in English (Step 3), and zero-shot transferred into the rest of languages (Step 4). We perform 5 independent finetuning runs with different random seeds, and report average results. During finetuning, we use a peak learning rate of 1e-5 for XNLI and PAWS-X, and 3e-5 for MLQA and XQuAD. Each uses a warmup ratio of 0.06 and linear decay, and is finetuned for 3 epochs.

\paragraph{Estimating FLOPs.} 
We compare training efficiency of different approaches using floating point operations (FLOPs). To calculate FLOPs, we estimate analytically using an adaptation of the formula from \citet{NarayananFlops2021}, detailed in \S\ref{app-sec:flops}. When doing so, we exclusively consider the cost of expanding the model to a new language (Step 2), which is the most significant in the cross-lingual lifelong learning setup that our work addresses.\footnote{While Step 1 can also be expensive, it is amortized over time: the initial model is trained only once, but extended to new languages many times. The cost of Step 1 is similar for \baselinelarge{} and \adhoc{}, as the overhead of the second head is small ($\sim$30.4 vs. $\sim$32.2 V100 days for a 12-layer model). \posthoc{} incurs extra cost from Step 1b, but this is relatively small compared to the cost of pretraining (see \S\ref{app-sec:flops}).} 
We also report NVIDIA V100 GPU training days as a more interpretable number, which we estimate analytically using an estimated throughput of 30~TFLOP/s, or 1 V100 day = 2.592~EFLOPs.

In some of our experiments, we are interested in estimating the training FLOPs required to achieve certain performance. However, this cannot be computed precisely, as we only have a limited number of intermediate checkpoints.\footnote{
We checkpoint every 5000 updates for \baselinesmall{}, \adhoc{}, \posthoc{}. As each step of \baselinelarge{} is more expensive, we checkpoint every 1000 updates for more fine-grained estimates. We save extra checkpoints for \adhoc{} and \posthoc{} at steps 1000, 2000, 3000 and 4000 for De, Fr, Es, and Zh, as these adapt rapidly for certain tasks. 
}
For that reason, we identify the checkpoints immediately before and after which the model first scores the desired performance, and use linear interpolation to estimate the step at which the exact score would have been hit. For instance, if \posthoc{} obtains an accuracy of 48\% at the 5,000 update checkpoint ($\sim$1.17 EFLOPs) and 52\% at the 10,000 update checkpoint ($\sim$2.34 EFLOPs), we estimate that the accuracy of 50\% was achieved at 7,500 steps ($\sim$1.76 EFLOPs).

\section{Main Results}

\subsection{Performance at Training Completion}
\label{sec:main}

\begin{table*}[t]

\begin{subfigure}{\textwidth}
\centering
\resizebox{\textwidth}{!}{
\addtolength{\tabcolsep}{-2.5pt}
\begin{tabular}{llllcclcccccccccccccca}
\toprule
&&
&& \multicolumn{2}{c}{Train cost}
&& \multicolumn{15}{c}{XNLI (acc)}
\\
\cmidrule{5-6}
\cmidrule{8-22}
&&&
& EFLOPs & days &
& ar & bg & de & el & es & fr & hi & ru & sw & th & tr & ur & vi & zh & \multicolumn{1}{c}{avg}
\\
\midrule
\multirow{2}{*}{Standard}
&& \baselinelarge{} && 54.1 & 20.9 && 
70.2 & 78.4 & 76.2 & 76.1 & 79.2 & 78.9 & 65.6 & 72.5 & 68.2 & 70.1 & 67.1 & 60.9 & 72.1 & 72.4 & 72.0\\
&& \baselinesmall{} && 21.1  & 8.1 && 
65.4 & 71.1 & 68.0 & 69.1 & 71.3 & 71.0 & 61.8 & 66.2 & 63.8 & 64.9 & 64.4 & 57.9 & 65.7 & 67.6 & 66.3 \\

\midrule

\multirow{2}{*}{Proposed}
&& \posthoc{} && 29.3 & 11.3 && 
70.1 & 77.8 & 75.7 & 75.5 & 78.5 & 78.2 & 63.9 & 72.2 & 67.5 & 70.2 & 64.8 & 59.2 & 70.8 & 72.0 & 71.2 \\
&& \adhoc{} && 21.1 & 8.1 &&
69.3 & 77.6 & 75.0 & 74.7 & 78.4 & 77.7 & 62.4 & 71.7 & 66.9 & 68.8 & 63.7 & 58.1 & 69.2 & 70.8 & 70.3 \\

\bottomrule
\end{tabular}
}
\caption{XNLI results}
\end{subfigure}

\vspace{10pt}

\begin{subfigure}{\textwidth}
\centering
\resizebox{0.87\textwidth}{!}{
	\addtolength{\tabcolsep}{-2.5pt}
	\begin{tabular}{llllcclccccccalcccca}
		\toprule
		&&
		&& \multicolumn{2}{c}{Train cost}
		&& \multicolumn{7}{c}{MLQA (F1)}
		&& \multicolumn{5}{c}{PAWS-X (acc)}
		\\
		\cmidrule{5-6}
		\cmidrule{8-14}
		\cmidrule{16-20}
		&&&
		& EFLOPs & days &
		& ar & de & es & hi & vi & zh & \multicolumn{1}{c}{avg} &
		& de & fr & es & zh & \multicolumn{1}{c}{avg}
		\\
		\midrule
		\multirow{2}{*}{Standard}
		&& \baselinelarge{} && 54.1 & 20.9 && 
                51.2 & 61.0 & 66.6 & 48.5 & 57.3 & 56.8 & 56.9 && 84.7 & 85.8 & 86.0 & 77.3 & 83.4 \\
		&& \baselinesmall{} && 21.1  & 8.1 && 
                46.0 & 54.6 & 59.8 & 43.1 & 53.2 & 50.8 & 51.2 &&  74.2 & 76.4 & 76.6 & 70.8 & 74.5 \\
		
		\midrule
		
		\multirow{2}{*}{Proposed}
		&& \posthoc{} && 29.3 & 11.3 &&  
                50.8 & 60.4 & 66.7 & 48.9 & 56.6 & 56.7 & 56.7 && 83.8 & 86.2 & 86.3 & 76.0 & 83.1  \\
		&& \adhoc{} && 21.1 & 8.1 && 
                50.8 & 60.1 & 66.0 & 46.4 & 57.2 & 55.6 & 56.0 && 84.4 & 85.1 & 86.7 & 77.9 & 83.5 \\
		\bottomrule
	\end{tabular}
}
\caption{MLQA and PAWS-X results}
\end{subfigure}

\caption{
\textbf{Performance at training completion.}
Both variants of our approach nearly match the performance of \baselinelarge{} at a substantially lower cost, while \baselinesmall{} significantly lags behind.
\textit{days}: V100 GPU days.
}
\label{tab:main}
\end{table*}

Table \ref{tab:main} reports performance at training completion (i.e., after 125,000 updates in Step 2).
As expected, \baselinelarge{} obtains the best results, but its training cost is also the highest.
In contrast, \adhoc{} requires nearly one third of the compute, while obtaining similar results. More concretely, it is marginally better on PAWS-X, while moderately (1-2 points) worse on MLQA and XNLI. Averaged over tasks, \adhoc{} retains 98.7\% of \baselinelarge{}'s performance\footnote{$(\frac{70.3}{72.0} + \frac{56.0}{56.9} + \frac{83.5}{83.4}) / 3 \approx 0.987$} at 39\% of its cost.
This validates the core hypothesis of our work---learning target language embeddings over the mini-model is almost as effective as learning them over the original model, while being significantly cheaper.

\posthoc{} follows a similar trend, retaining 99.3\% of \baselinelarge{}'s performance at nearly half of its cost. This shows that mini-models do not need to be trained from scratch, but one can take any existing English model and build it's corresponding mini-model post-hoc.

\baselinesmall{} performs substantially worse than our proposed approach. \baselinesmall{} has the same training cost as \adhoc{}, but is 4.0 points worse on XNLI, 4.8 points worse on MLQA, and 9.0 points worse on PAWS-X. This shows that our idea of having two aligned models---a shallow one for efficient adaptation and a deep one for best performance at test time---is critical, as using a shallow model both for adaptation and inference performs considerably worse.

\subsection{GPU days to Near-Maximal Performance}
\label{sec:near-max}

While we previously compared approaches at training completion, one can also apply early stopping, sacrificing some performance to gain on efficiency. This also allows to compare different approaches head-to-head according to the compute they require to achieve a given score---assuming we stop training as soon as the desired performance is hit. To that end, we fix our target score as 95\% of the performance obtained by \baselinelarge{} at the end of training, which we call \textit{near-maximal performance}.\footnote{For instance, \baselinelarge{} obtains 70.2 accuracy on Arabic XNLI at training completion, so we set the target performance for Arabic XNLI at $70.2*0.95=66.7$. Note that this is also the target performance we use for \adhoc{} and \posthoc{}, even if their score at training completion is lower.} Results are in Table \ref{tab:95}, and average speedup of our approach over standard adaptation is in Figure \ref{fig:speedup}.\footnote{
\textit{Speedup} is the ratio of V100 days to \textit{near-maximal performance} between two methods. For instance, if \baselinelarge{} requires 2.5 V100 days to achieve near-maximal performance in Arabic XNLI and \adhoc{} requires 1.0, the resulting speedup is $2.5/1.0\approx2.5$. Note that Figure \ref{fig:speedup} reports the average ratio across all languages, which is not the same as the ratio of the average V100 days across all languages.
}

Overall, \adhoc{} does best: when per-language speedup is averaged across languages, we see that it requires about half to one-third the compute of \baselinelarge{} to achieve the same performance in all tasks.
\posthoc{} has more modest speedups, but is still substantially faster than standard adaptation to hit the desired performance. This shows that, if possible, it is preferable to pretrain mini-models jointly with the primary model, but our approach can also bring substantial speedups when starting with an existing pretrained model.

It is also remarkable that there is a considerable variance across tasks. In particular, all approaches require substantially less compute to achieve the target performance in PAWS-X when compared to XNLI and MLQA. The relative speedup of mini-model adaptation is also considerably higher on PAWS-X.
We also observe a high variance across languages, which we analyze in more detail in \S\ref{subsec:variance}.

\begin{table*}[t]
\centering
\resizebox{\textwidth}{!}{
\addtolength{\tabcolsep}{-3.5pt}
\begin{tabular}{llccccccccccccccalccccccalcccca}
\toprule
&& \multicolumn{15}{c}{XNLI}
&& \multicolumn{7}{c}{MLQA}
&& \multicolumn{5}{c}{PAWS-X}
\\
\cmidrule{3-17}
\cmidrule{19-25}
\cmidrule{27-31}
&
& ar & bg & de & el & es & fr & hi & ru & sw & th & tr & ur & vi & zh & \multicolumn{1}{c}{avg} &
& ar & de & es & hi & vi & zh & \multicolumn{1}{c}{avg} &
& de & fr & es & zh & \multicolumn{1}{c}{avg}
\\
\midrule
\baselinelarge{} &
& 2.5 & 1.2 & 1.1 & 1.5 & 0.8 & 0.8 & 3.2 & 1.8 & 1.4 & 1.8 & \textbf{2.3} & 8.3 & 1.0 & 1.7 & 2.1 & 
& 2.6 & 1.1 & 0.8 & 3.4 & 1.2 & 1.8 & 1.8 &  
& 0.7 & 0.6 & 0.6 & 0.7 & 0.6  
\\
\midrule
\posthoc{} &
& 1.7 & 0.9 & 0.8 & 0.9 & 0.4 & 0.4 & \textbf{2.5} & 1.3 & 1.1 & 1.3 & 3.0 & 6.5 & 0.8 & 1.1 & \textbf{1.6} & 
& 1.7 & 0.8 & 0.5 & \textbf{1.7} & 0.9 & 1.2 & \textbf{1.1}  & 
& 0.3 & 0.3 & 0.3 & 0.4 & 0.3 
\\
\adhoc{} &
& \textbf{1.0} & \textbf{0.5} & \textbf{0.5} & \textbf{0.6} & \textbf{0.3} & \textbf{0.3} & 5.9 & \textbf{0.6} & \textbf{0.6} & \textbf{0.7} & 5.3 & \textbf{5.4} & \textbf{0.6} & \textbf{0.8} & \textbf{1.6} & 
& \textbf{1.0} & \textbf{0.6} & \textbf{0.4} & 5.3 & \textbf{0.5} & \textbf{0.9} & 1.5 & 
& \textbf{0.2} & \textbf{0.2} & \textbf{0.2} & \textbf{0.2} & \textbf{0.2}  
\\
\bottomrule
\end{tabular}
}
\caption{
\textbf{Estimated V100 training days to achieve near-maximal performance.}
Near-maximal performance is defined as 95\% of the score of \baselinelarge{} at training completion. $\downarrow$ is better. \baselinesmall{} never achieves near-maximal performance, except on XNLI for Turkish (1.7 days) and Urdu (6.4 days).
}
\label{tab:95}
\end{table*}

\begin{figure*}[ht]
\centering
\makebox[\textwidth]{%
\includegraphics[width=0.33\textwidth,trim={0.8cm 0.1cm 1.3cm 0.5cm},clip]{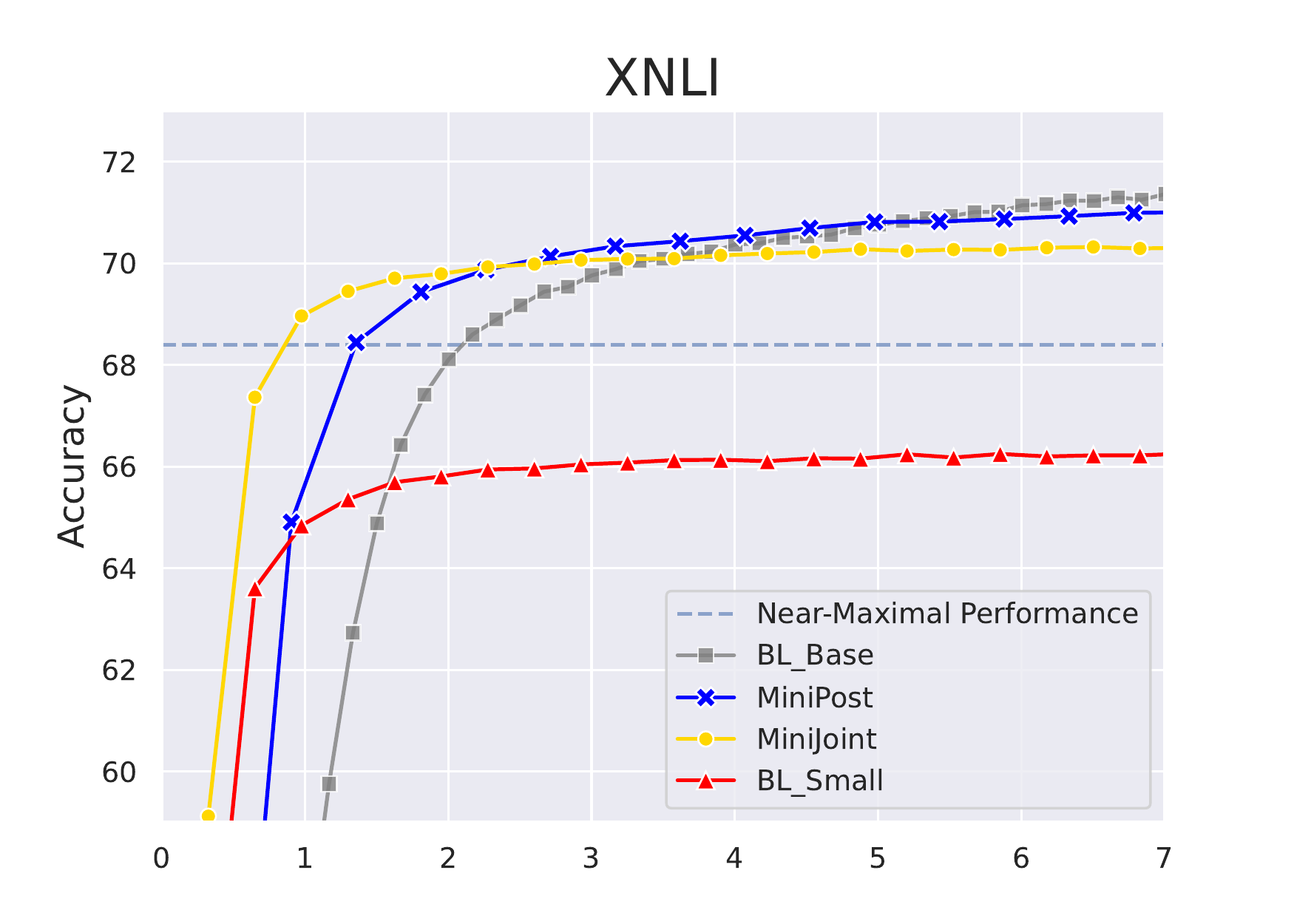}%
\includegraphics[width=0.33\textwidth,trim={0.8cm 0.1cm 1.3cm 0.5cm},clip]{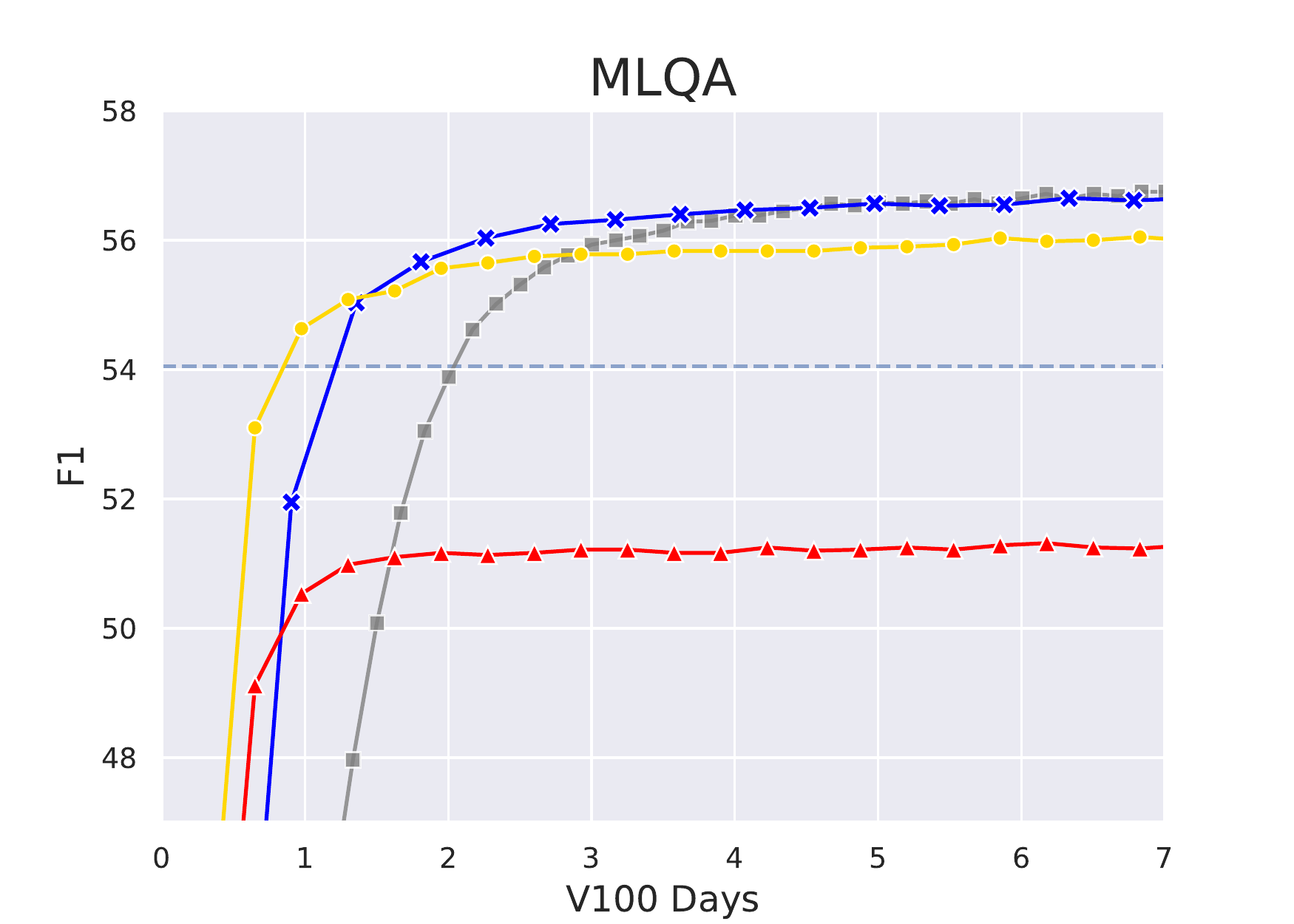}%
\includegraphics[width=0.33\textwidth,trim={0.8cm 0.1cm 1.3cm 0.5cm},clip]{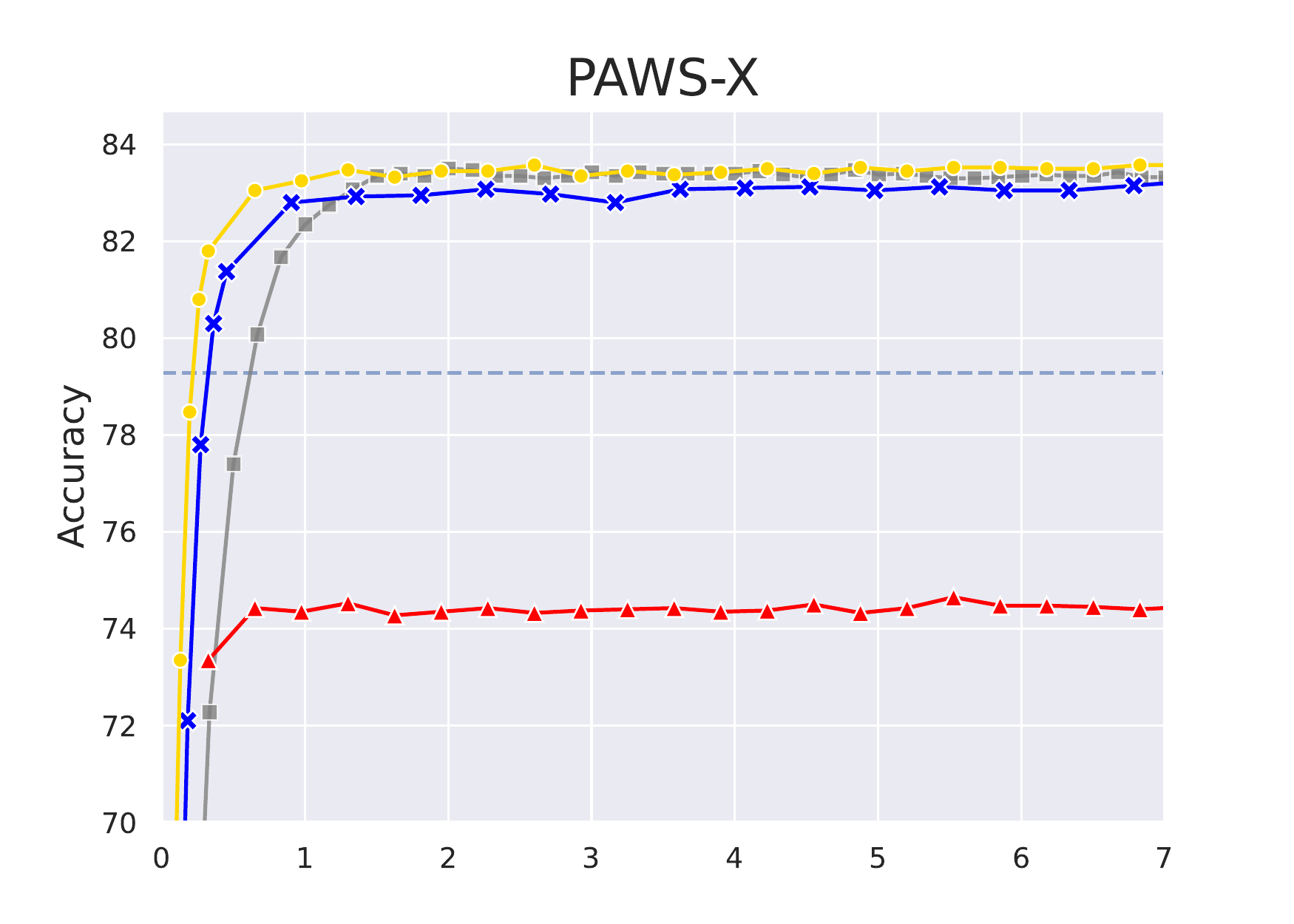}%
}\\
\caption{\textbf{Training curve through the first GPU-week.} We report XNLI and PAWS-X accuracy and MLQA F1.}
\label{fig:curves}
\end{figure*}

\section{Analysis}

\subsection{Training Curves}

We visualize the training curves of the different approaches in Figure \ref{fig:curves}. Consistent with our previous findings, we observe that \adhoc{} is usually the leftmost curve---signifying the most rapid adaptation---at the cost of a slightly lower final score. In contrast, \baselinelarge{} is by far the slowest system approaching its peak performance, while \baselinesmall{} gets stuck at a poor performance compared to other approaches. Finally, we find that all methods adapt rapidly in PAWS-X, which suggests that this tasks might be easier than the others.

\subsection{Mini-Model Depth}
\label{sec:layerwise}

We recall that the mini-model in \adhoc{} has 4 layers, whereas the one in \posthoc{} has 6 (the bottom 4 taken from the primary model + 2 additional ones trained on top). We made this decision early in the development process based on preliminary experiments. In this section, we more systematically study the effect of mini-model depth on efficiency and performance. To that end, we build models with the same architecture as \adhoc{}, but placing the secondary MLM head after layers 2, 6, 10, or 12.\footnote{Attaching after layer 12 means that both heads are at the final layer, making it virtually equivalent to \baselinelarge{}.} We experiment with Arabic, German and Turkish due to compute constraints.

\begin{figure}[t]
\centering
\includegraphics[width=1\linewidth]{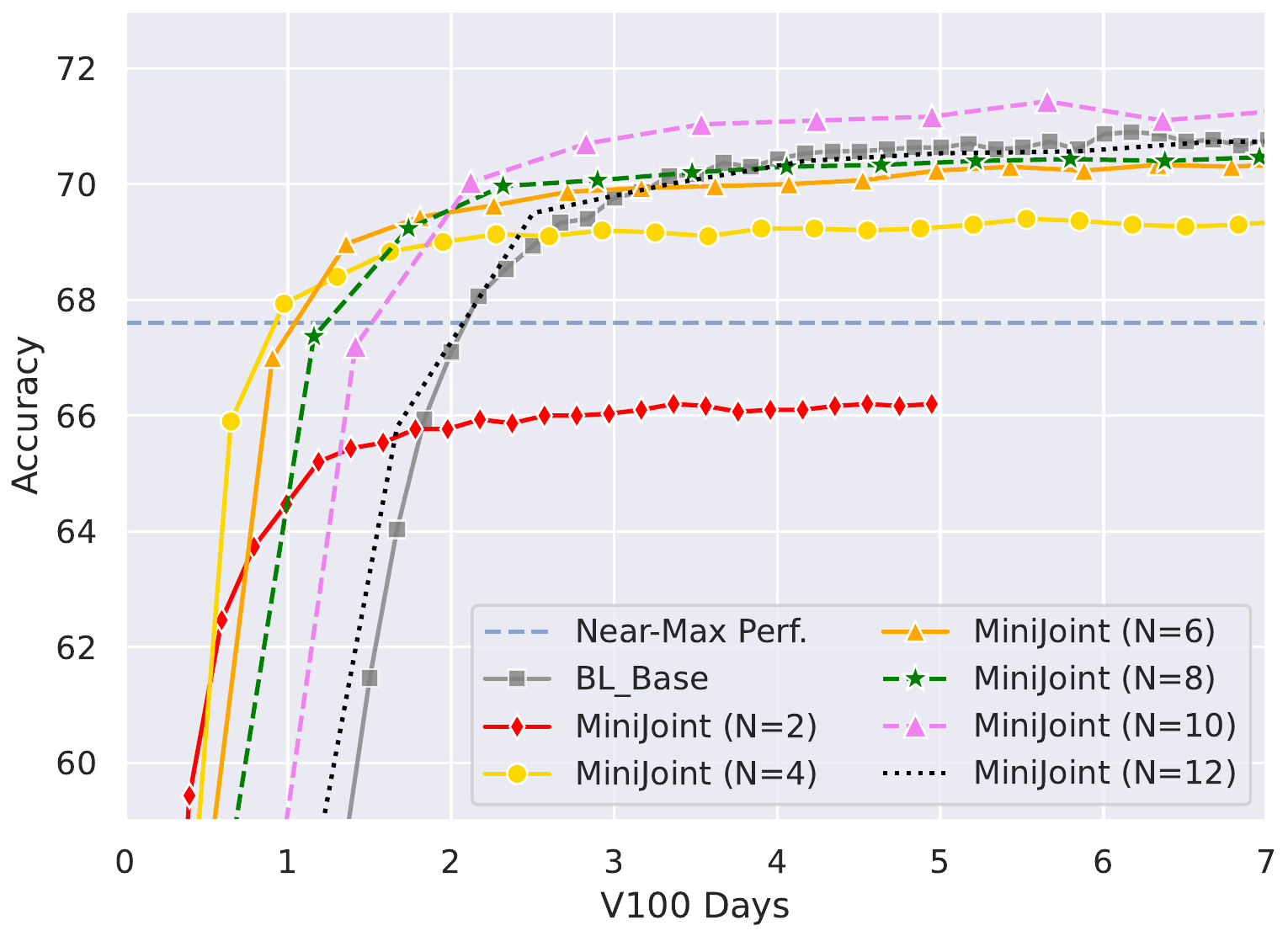}
\caption{\textbf{XNLI training curve for \adhoc{} with secondary head attached at varying layers.} Results are averaged over Arabic, German and Turkish. Final performance is in Table \ref{tab:layerwise-final}.}
\label{fig:layerwise-xnli-avg}
\end{figure}

Figure \ref{fig:layerwise-xnli-avg} shows the XNLI training curve averaged over 3 languages. We see more rapid adaptation with shallower attachment of the second head, at a cost to final performance.
\S\ref{sec:app-layerwise} shows curves for PAWS-X, MLQA, and XQuAD. For PAWS-X, high performance was rapidly achieved by all models. End-of-training results are in Table \ref{tab:layerwise-final}.

Table \ref{tab:layerwise-interp-95} reports estimated V100 days to achieve near-maximal performance as defined in \S\ref{sec:near-max}, and upper and lower estimates are in \S\ref{sec:app-layerwise}. We find that the optimal depth of the mini-model is largely language-dependent. Specifically, Arabic and Turkish never hit the target performance with 2 layers, whereas German does so quickly. For Arabic, 4 layers provides the most rapid adaptation, while Turkish requires at least 6. This suggests that it is critical to have some minimum number of layers to achieve good performance, which varies from language to language. But, as long as this minimum is met, shallower mini-models are generally more efficient.

\begin{table}[t]
\centering\small
\begin{tabular}{lrlrrrrr}
\toprule
 &
  \multicolumn{1}{l}{\textbf{Layer:}} &
  \multicolumn{1}{c}{\textbf{2}} &
  \multicolumn{1}{c}{\textbf{4}} &
  \multicolumn{1}{c}{\textbf{6}} &
  \multicolumn{1}{c}{\textbf{8}} &
  \multicolumn{1}{c}{\textbf{10}} &
  \multicolumn{1}{c}{\textbf{12}} \\ \midrule
 &
  XNLI &
  \multicolumn{1}{r}{
  \cellcolor[HTML]{65FF00}{0.4}} &
  \cellcolor[HTML]{93FF00}{0.5} &
  \cellcolor[HTML]{D9FF00}{0.7} &
  \cellcolor[HTML]{FEF705}{1.0} &
  \cellcolor[HTML]{FCEE0B}{1.1} &
  \cellcolor[HTML]{F8DB18}{1.4} \\
 &
  MLQA &
  \multicolumn{1}{r}{
  \cellcolor[HTML]{D0FF00}{0.7}} &
  \cellcolor[HTML]{A7FF00}{0.6} &
  \cellcolor[HTML]{C1FF00}{0.6} &
  \cellcolor[HTML]{FDF308}{1.1} &
  \cellcolor[HTML]{FDF308}{1.0} &
  \cellcolor[HTML]{F8DB18}{1.4} \\
\multirow{-3}{*}{\textbf{de}} &
  PAWS &
  \multicolumn{1}{r}{
  \cellcolor[HTML]{36FF00}{0.2}} &
  \cellcolor[HTML]{41FF00}{0.2} &
  \cellcolor[HTML]{83FF00}{0.4} &
  \cellcolor[HTML]{ABFF00}{0.5} &
  \cellcolor[HTML]{D6FF00}{0.7} &
  \cellcolor[HTML]{FDF308}{1.0} \\ \midrule
 &
  XNLI &
  \cellcolor[HTML]{E34747}{$\infty$} &
  \cellcolor[HTML]{FEFA03}{1.0} &
  \cellcolor[HTML]{FEFA03}{0.9} &
  \cellcolor[HTML]{FAE511}{1.3} &
  \cellcolor[HTML]{F3C427}{1.8} &
  \cellcolor[HTML]{EDA33D}{2.4} \\
\multirow{-2}{*}{\textbf{ar}} &
  MLQA &
  \cellcolor[HTML]{E34747}{$\infty$} &
  \cellcolor[HTML]{FEF705}{1.0} &
  \cellcolor[HTML]{FDF308}{1.1} &
  \cellcolor[HTML]{EFAE36}{2.2} &
  \cellcolor[HTML]{F1B730}{2.1} &
  \cellcolor[HTML]{EC9D41}{2.5} \\ \midrule
\textbf{tr} & XNLI &
  \cellcolor[HTML]{E34747}{$\infty$} &
  \cellcolor[HTML]{E06666}{5.3} &
  \cellcolor[HTML]{F8D919}{1.5} &
  \cellcolor[HTML]{F8D919}{1.5} &
  \cellcolor[HTML]{F8D919}{1.5} &
  \cellcolor[HTML]{F6CF20}{1.6} \\
  \bottomrule
\end{tabular}
\caption{\textbf{Estimated V100 days to near-maximal performance (see \S\ref{sec:near-max}) for \adhoc{} with secondary head attached at varying layers.} $\downarrow$ better. $\infty$: never hit target performance.}
\label{tab:layerwise-interp-95}
\end{table}

\subsection{English Performance} \label{subsec:english}
While all of our results so far correspond to the target languages, we next look into the source language performance. As described in \S\ref{subsec:method}, \posthoc{} uses \baselinelarge{} as the primary model, so their English performance is exactly the same. However, \adhoc{} jointly pretrains the primary model and its aligned mini-model. To understand the effect of the joint pretraining on the monolingual quality of the model, we compare the full \adhoc{} model and its corresponding mini-model with \baselinelarge{} and \baselinesmall{}. As shown in Table \ref{tab:vanilla-ft}, we find that dual-head training does not damage performance: the full \adhoc{} model performs on-par with the 12-layer baseline, and the 4-layer extracted mini-model performs on-par with the 4-layer baseline.

\begin{table}[t]
\centering
\begin{tabular}{@{}lc@{}}
\toprule
\baselinelarge{}      & 86.4 \\
\baselinesmall{}      & 79.6 \\
\adhoc{} (full) & 86.2 \\
\adhoc{} (mini-model) & 79.2 \\
\bottomrule
\end{tabular}
\caption{\textbf{English XNLI accuracy.} \S\ref{subsec:english} for details.}
\label{tab:vanilla-ft}
\end{table}

\subsection{Variance Across Languages} \label{subsec:variance}
While we obtain strong results across the board, there are 3 languages that prove challenging: Hindi, Turkish and Urdu. As shown in Table \ref{tab:95}, \adhoc{} takes more than 5 V100 days to achieve near-maximal performance on XNLI for these languages, whereas the rest of the languages require at most 1 day. As seen in \S\ref{sec:layerwise} this can be mitigated by using a deeper mini-model in the case of Turkish. However, we observe that even \baselinelarge{} struggles with Urdu and, to a lesser extent, Hindi. This suggests that there is something making these languages particularly challenging for cross-lingual adaptation, affecting not only our method but also the standard approach from \citet{artetxe-etal-2020-cross}.

One hypothesis is that this is due to the high linguistic distance between these languages and English. In Table \ref{tab:data}, these are the languages that are the most syntactically distant from English according to \textit{lang2vec},\footnote{\url{https://github.com/antonisa/lang2vec}} and the only ones with a pure SOV word order. This is also consistent with German, Spanish and French---the 3 languages that are the closest to English---generally obtaining the fastest adaptation times. In the future, we would like to explore starting with a multilingual model covering a few diverse languages akin to \citet{pfeiffer-etal-2022-lifting}, which could facilitate adapting to languages that are distant from English but might share features with some of the other languages.

Another potential factor is that Hindi, Turkish and Urdu, along with Swahili, have the smallest training corpora. However, despite having the smallest training corpus with only 1.7GB---$\sim$1/3 the size of Urdu and $\sim$1/12 of Hindi and Turkish---Swahili exceeds the aforementioned three on both adaptation speed and raw performance on XNLI. Exploring the impact of corpus size was outside of the scope of this work, but we believe that this is an interesting question to address in future work.

\section{Related Work}

\paragraph{Multilinguality in NLP.}
One way to create a LM for a particular language is to collect enough data and train from scratch \cite[e.g.][]{martin-etal-2020-camembert, de2019bertje,chan-etal-2020-germans}. For the majority of languages, however, not enough data exists to train a high-quality model from scratch. 
Alternatively, one may pretrain a multilingual model on unlabeled data from many languages, which can then be finetuned on labeled data for zero-shot cross-lingual transfer \cite[e.g.][]{devlin-etal-2019-bert, conneau2019cross, conneau-etal-2020-unsupervised}.
Multilingual LMs are not without challenges; they are large and expensive to train, suffer from the \textit{curse of multilinguality}, low-resource language performance can lag due to underrepresentation in the training corpus, and they cannot benefit from language-specific tokenization \cite[][for a survey]{conneau2019cross, wu-dredze-2020-languages,rust-etal-2021-good,doddapaneni2021primer}. 
Furthermore, not all languages are created alike in multilingual models; \citet{muller-etal-2021-unseen} find that some ``easy'' languages perform well in mBERT out-of-the-box and others are successfully after finetuning with monolingual data, some ``hard'' languages perform poorly in mBERT even after tuning. 
Alternatively, one may adapt a pretrained model by finetuning, adding language- or domain-specific adapters \cite[e.g.][]{rebuffi2017learning, houlsby2019parameter, pfeiffer-etal-2022-lifting}, retraining the lexical embedding layer \cite{tran2020english,artetxe-etal-2020-cross,de-vries-nissim-2021-good}, or translating the train, finetuning, or test set \cite[e.g.][]{wang-etal-2022-expanding}.

\paragraph{Efficient Adaptation of Language Models.}
Adapters are a \textbf{parameter-efficient} way to extend LMs by training a small number of parameters that can be swapped-in for on-the-fly adaptation at test time as opposed to needing to store full separate models per task or language.  
\citet{pfeiffer-etal-2020-mad} train small stackable language- and task-specific adapters with respect to a frozen transformer body that is shared between all languages and tasks, allowing simple and quick cross-lingual transfer at test-time.  
\citet{bapna-firat-2019-simple} inject adapter layers into a neural machine translation (NMT) model for domain adaptation to obviate the need for full-model finetuning, and use language-specific adapters for high-resource languages to recover from catastrophic forgetting during multilingual NMT training.
\citet{alabi-etal-2022-adapting} argue that their finetuned mBERT for 17 African languages is parameter efficient because they maintain high-performance with a single model rather than requiring separate models per language. 
Like \citet{abdaoui-etal-2020-load}, they reduce model size by removing vocabulary tokens not needed for target languages. LoRa adds small trainable matrices corresponding to low-rank decompositions of a weight updates within transformer attention, allowing rapid updates during finetuning \citep{hu2022lora}. Prefix-tuning methods are also parameter-efficient \citep{li-liang-2021-prefix, liu2021gpt}.

\textbf{Compute-efficient} methods aim reduce the computation (FLOPs or wall-time) required to train a model.
Several authors developed vocabulary adaptation methods which reduce the need to extensively finetune a model or train from scratch \cite[e.g.][]{chronopoulou-etal-2020-reusing,sachidananda-etal-2021-efficient}. 
Though \citet{wang-etal-2020-extending} continued-train mBERT with an extended vocabulary for a new language, convergence is faster than with a bilingual BERT model trained from scratch.
\citet{kocmi-bojar-2020-efficiently}'s vocabulary adaptation method improves time-to-convergence of a NMT system adapted to a new language.
While \citet{de-vries-nissim-2021-good} learn a new lexical embedding layer on top of GPT-2, which is computationally expensive, they employ engineering strategies to decrease training time, such as 16-bit mixed precision training, reduced window size, and maximum batch size with gradient accumulation.  
Though they must backpropogate through the entire model during embedding layer relearning, training stabilizes quickly. 
They adapt larger models by initializing the embedding layer using transformations of embeddings developed on smaller models, noting that the better initialization speeds training.

\paragraph{Variance across languages.} Prior work observes similar variation between languages in LM adaptation.
When adapting BERT, \citet{tran2020english} see that Hindi showed the slowest growth and lowest final XNLI score of six assessed languages, acknowledging word-order differences. 
Several authors see performance lags on NLP benchmarks for SOV languages when probing large multilingual models \cite[][for a review]{doddapaneni2021primer}. 
\citet{pires-etal-2019-multilingual} find that zero-shot part-of-speech tagging is best when the model has been finetuned on a language that shares word order with the target language. 
\citet{limisiewicz-etal-2020-universal} attribute the disparity to underrepresentation of SOV languages in the training corpus. 

\section{Conclusion and Future Work}

Our work shows that it is possible to extend pretrained models to new languages using only a fraction of their parameters. We achieve this by learning a new embedding layer over a shallow \textit{mini-model} aligned with the primary model. We explore two approaches to learn mini-models: \adhoc{} augments a transformer with a second MLM head during pretraining, adapting with an average 2.3x speedup over the standard method from \citet{artetxe-etal-2020-cross}, and \posthoc{} builds a mini-model by extracting a small number of layers from a pretrained model, providing an average 1.6x speedup.

Our analysis reveals that shallower mini-models converge faster but plateau at lower performance. As such, one might explore combining multiple mini-models of different depths, using the shallowest at the beginning of cross-lingual adaptation, and then deeper ones as training progresses. One could add multiple MLM heads to a \adhoc{} model and train all simultaneously to facilitate this.

We would also like to explore applications of mini-model adaptation beyond the multilingual scenario.
In particular, by adapting rapidly on models significantly smaller than the base model used for inference, \adhoc{}/\posthoc{} might be used to finetune large LMs on modest hardware.  
This could allow for a new paradigm whereby one shares a small model for adaptation while keeping a large aligned model private behind an API. Clients could then learn parameters for their task on the small model, which are later plugged into the large model for better performance. Shortly after us, \citet{xiao2023offsite} proposed \textit{Offsite-Tuning}, an adaptation method similar to ours but motivated by privacy.

\section*{Limitations}
Our study is limited to the adaptation of MLMs to new languages. While we believe that our proposed approach could also be applied more broadly (e.g., autoregressive models instead of MLMs, or adapting to new downstream tasks instead of new languages), further experiments are necessary to empirically verify this. In addition, we observe a considerable variance across languages (\S\ref{subsec:variance}), the reasons for which are not entirely clear.
Ideally, we would have a broader set of languages to better study this, as our language set is limited and skewed towards the Indo-European family. Finally, we average results over 5 finetuning runs, but computational restrictions prevented us from also averaging over multiple pretraining runs. As discussed in \S\ref{app:pretraining-seed}, we observed a non-negligible variance over pretraining runs in a preliminary experiment, but a more systematic exploration is necessary to better understand its impact.

\section*{Acknowledgements} The authors would like to thank Patrick Littell for helpful discussions about \textit{lang2vec}, along with Philipp Koehn, Elina Baral, Sophia Hager, and Mathias Unberath for helpful discussions and feedback. 

\bibliography{anthology,custom}
\bibliographystyle{acl_natbib}

\clearpage
\appendix

\setcounter{table}{0}
\setcounter{figure}{0}
\renewcommand{\thetable}{A\arabic{table}}
\renewcommand{\thefigure}{A\arabic{figure}}
\renewcommand{\theequation}{A\arabic{equation}}

\section{Appendix}
\label{sec:appendix}

\subsection{Floating Point Operations (FLOPs)}
\label{app-sec:flops}
We estimate total FLOPs for training using the formula from \citet{NarayananFlops2021}, amended for RoBERTa without activation recomputation. Like the authors, we omit calculations over biases, activation functions, softmax, and other minor costs.
Assume hidden size $h$, vocabulary size $V$, number of layers $l$, token mask probability $p$, sequence length $s$, batch size $B$, and total training updates $U$, the total FLOPs during training are:
\begin{equation}
\label{eq:flops}
72UBslh^2(1 + \frac{s}{6h} + \frac{p}{12l} + \frac{pV}{12hl})
\end{equation}

\paragraph{Derivation} Recall that multiplying $A \in \mathbb{R}^{m \times n}$ by $B \in \mathbb{R}^{n \times p}$ requires $2mnp$ FLOPs.
Each transformer layer consists of a multi-head self-attention block and a linear projection.  
The attention block has four weight matrices $W_q, W_k, W_v, W_o \in \mathbb{R}^{h \times h}$.\footnote{We demonstrate the calculation over one head, as using more heads results in the same FLOPs calculation.}
The input $x \in \mathbb{R}^{s \times h}$ is projected with $W_q, W_k$ and $W_v$, requiring $2sh^2$ FLOPs each:
\[
Q = xW_q \qquad
K = xW_k \qquad
V = xW_v
\]
Self-attention followed by output projection is: 
\[
(\text{softmax}(\frac{QK^T}{\sqrt{h}})V)W_O
\]
Multiplying $QK^T$ and multiplying the result by $V$ both require $2hs^2$ FLOPs. Multiplying with $W_O$ costs $2sh^2$ FLOPs.  
In sum, there are $8sh^2 + 4hs^2$ FLOPs to compute the forward pass of the attention block. 
The output of the attention block ($x \in \mathbb{R}^{s \times h}$) is then passed through two linear layers: $F_0 \in \mathbb{R}^{h \times 4h}$ and $F_1 \in \mathbb{R}^{4h \times h}$. 
These multiplications cost $8sh^2$ FLOPs each, so total FLOPs per layer is: 
\[\text{FLOP}_\text{layer} = 24sh^2 + 4hs^2\] 
The output $x \in \mathbb{R}^{s \times h}$ passes through the MLM head: a dense layer of size $\mathbb{R}^{h \times h}$ for $2sh^2$ FLOPs, and an output projection of size $\mathbb{R}^{h \times V}$ that costs: 
\[\text{FLOP}_\text{outproj} = 2shV\] 
Only masked tokens are passed through MLM head, so the total flops in the LM head is 
\[\text{FLOP}_\text{lm} = p(2sh^2 + 2shV)\]

In sum, the total estimated FLOPs for a forward pass of RoBERTa with a batch size of 1 is:
\begin{equation}
\label{eq:fwd}
\begin{split}
l (\text{FLOP}_\text{layer}) + \text{FLOP}_\text{lm} \\
= l(24sh^2 + 4hs^2) + p(2sh^2 + 2shV)
\end{split}
\end{equation}
To account for the backward pass, one typically triples the forward pass FLOPs.  
This is because (1) to backpropogate the error, one calculates the partial derivatives of the loss with respect to the input (activations): $\frac{\partial{\delta}}{\partial{a}}$, and (2) to make a weight update, one first must calculate the partial derivatives with respect to the weights: $\frac{\partial{\delta}}{\partial{w}}$.  
Calculating each partial derivative requires the same number of FLOPs as the forward pass, meaning that the backward pass is doubly as expensive.\footnote{One typically does not add in the cost of the weight update, because this is relatively small.} 
Tripling Equation \ref{eq:fwd} to account for the backward pass, multiplying by batch size and total updates, and reducing gives Equation \ref{eq:flops} for full pretraining.

Adaptation requires an amended equation for the backward pass because layers are frozen (Step 2: $\text{L}_\text{trg}$ embedding training). 
The trainable embeddings are tied to the output projection layer in the MLM head: thus, trainable input embeddings are passed through frozen layers, which passes through the MLM head consisting of a frozen dense layer and \textit{trainable} output projection.  
To backpropogate the error to the embeddings, we must (1) calculate $\frac{\partial{\delta}}{\partial{a}}$ for the entire model, requiring the same number of FLOPs as the forward pass.\footnote{Some additional backward computation is required here, but we make this simplification.}  
Because the MLM head's output projection layer is also trainable, we also calculate $\frac{\partial{\delta}}{\partial{w}}$ here on the backward pass.  In total, this gives the below equation for Step 2, after multiplying for batch size and total updates:
\begin{equation}
\begin{split}
UB (2l\text{FLOP}_\text{layer} + 2 \text{FLOP}_\text{lm} + p\text{FLOP}_\text{outproj}) \\
= 48UBslh^2(1 + \frac{s}{6h} + \frac{p}{12l} + \frac{pV}{8hl})
\end{split}
\end{equation}

Thus, adaptation with 4 layers requires $\sim$21.1~EFLOPs versus $\sim$29.3~EFLOPs during pretraining.  For 12 layers, adaptation requires $\sim$54.1~EFLOPs versus $\sim$78.8 in pretraining. 

\pagebreak
\paragraph{\posthoc{} FLOPs in Step 1b}

Step 1b of \posthoc{} builds small mini-model with embeddings and first $l_f$ layers frozen.  
These frozen layers do not require the backward pass.
Furthermore, the frozen LM head does not require calculating $\frac{\partial{\delta}}{\partial{w}}$, only $\frac{\partial{\delta}}{\partial{a}}$.  Of the trainable layers, each require both $\frac{\partial{\delta}}{\partial{a}}$ and $\frac{\partial{\delta}}{\partial{w}}$, except the first trainable layer which only needs $\frac{\partial{\delta}}{\partial{w}}$ (because it does not pass back the error).  Given trainable layers $l_t$, the total cost for creating the mini-model in \posthoc{} is:
\begin{small}
\begin{gather}
     = UB (l (\text{FLOP}_\text{layer}) + 2\text{FLOP}_\text{lm} + (2 l_t - 1)\text{FLOP}_\text{layer}) \nonumber \\ 
     = UB ((l + 2 l_t - 1) \text{FLOP}_\text{layer} + 2\text{FLOP}_\text{lm}) \nonumber \\
     = UB((l + 2l_t - 1) (24sh^2 + 4hs^2) + 4psh^2 + 4pshV)
\end{gather}
\end{small}

Concretely, the cost of training a 6-layer mini-model in this work is $\sim$21.6 EFLOPs.  In comparison, pretraining the vanilla 12-layer RoBERTa base model requires $\sim$78.8 EFLOPs. 

\subsection{XQuAD}
\label{sec:app-xquad}
\begin{figure}[htb]
\centering
\includegraphics[width=1\linewidth, trim={0 0.1cm 0 0},clip]{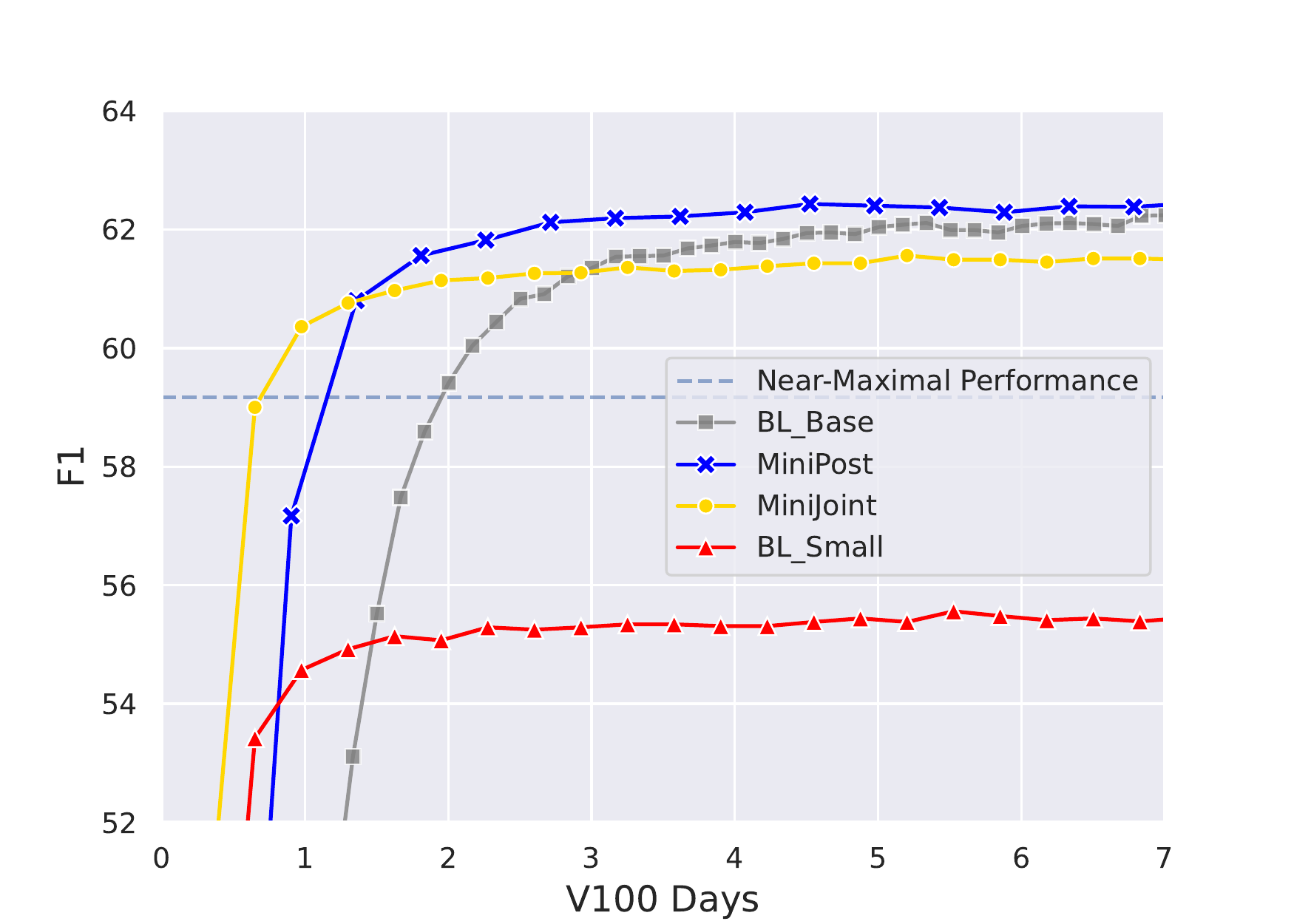}
\caption{\textbf{XQuAD performance in first GPU-week.}} 
\label{fig:app_xnli_curve}
\end{figure}

The Cross-lingual Question Answering Dataset \cite[XQuAD;][]{artetxe-etal-2020-cross} covers a more extensive set of languages than MLQA.  
We evaluate the same models tuned for QA in the main body of the paper on XQuAD.
Final F1 and V100 days to achieve near-maximal performance are in Tables \ref{tab:app_xquad_days} and \ref{tab:app_xquad_main}.
We also show the growth curve for F1 through the first V100-week in Figure \ref{fig:app_xnli_curve}.

\begin{table}[!hbt]                             
\centering                                    
\resizebox{\linewidth}{!}{ 
\addtolength{\tabcolsep}{-3.5pt}              
\begin{tabular}{@{}llcccccccccca}              
\toprule
&& \multicolumn{11}{c}{XQuAD}
\\
\cmidrule{3-13}
&
& ar & de & el & es & hi & ru & th & tr & vi & zh & \multicolumn{1}{c}{avg}       
\\
\midrule
\baselinelarge{}  &
& 2.3 & 1.1 & 1.7 & 0.8 & 3.3 & 1.9 & 2.1 & 2.2 & 1.2 & 1.5 & 1.8
\\
\midrule
\posthoc{} &                     
& 1.4 & 0.8 & 1.0 & \textbf{0.4} & \textbf{1.3} & 1.3 & 1.4 & \textbf{1.2} & 0.8 & 1.0 & 1.1
\\
\adhoc{} &  
& \textbf{0.8} & \textbf{0.4} & \textbf{0.6} & 0.5 & 3.1 & \textbf{0.6} & \textbf{0.6} & $\infty$ & \textbf{0.5} & \textbf{0.6} & \textbf{0.9}*                              
\\
    \bottomrule
    \end{tabular}
    }
\caption{
\textbf{Estimated V100 training days to achieve near-maximal performance (see \S\ref{sec:near-max}) on XQuAD.} $\infty$: never hit target performance. \baselinesmall{} never achieves near-maximal performance. $*$excludes Turkish, which never hit near-maximal performance.         
}
  \label{tab:app_xquad_days}
\end{table} 

\begin{table*}[!b]
  \centering
  \resizebox{\textwidth}{!}{                  
  \addtolength{\tabcolsep}{-2.5pt}            
    \begin{tabular}{llllcclcccccccccca}           
    \toprule
    &&
    && \multicolumn{2}{c}{Train cost}         
    && \multicolumn{11}{c}{XQuAD (acc)}
    \\
    \cmidrule{5-6}                            
    \cmidrule{8-18} 
    &&&                                       
 & EFLOPs & days &                            
 & ar & de & el & es & hi & ru & th & tr & vi & zh & \multicolumn{1}{c}{avg}
 \\   
  \midrule                                    
\multirow{2}{*}{Standard}                     
&& \baselinelarge{} && 54.1  & 20.9 && 
52.4 & 69.2 & 70.0 & 74.8 & 53.8 & 68.7 & 57.1 & 57.2 & 65.9 & 53.7 & 62.3 \\ 
&& \baselinesmall{} && 21.1  & 8.1 && 
48.2 & 61.2 & 63.3 & 65.8 & 46.2 & 61.0 & 50.4 & 52.1 & 60.3 & 46.2 & 55.5 \\
\midrule
\multirow{2}{*}{Proposed}
&& \posthoc{} && 29.3 & 11.3 && 
52.4 & 68.8 & 69.9 & 75.1 & 55.3 & 68.2 & 56.9 & 58.6 & 65.7 & 53.5 & 62.4 \\        
&& \adhoc{} && 21.1  & 8.1 && 
53.8 & 70.1 & 69.9 & 73.7 & 51.8 & 68.6 & 56.5 & 53.2 & 64.9 & 52.5 & 61.5 \\         
    \bottomrule                               
    \end{tabular}
    }
  \caption{
\textbf{XQuAD performance at training completion.}            
Both variants of our approach nearly match the performance of \baselinelarge{} at a substantially lower cost, while \baselinesmall{} significantly lags behind.
}   
  \label{tab:app_xquad_main}                       
\end{table*} 

\subsection{Mini-Model Depth: MLQA, PAWS-X, and XQuAD}
\label{sec:app-layerwise}
\begin{figure}
    \begin{subfigure}{\linewidth}
    \centering
    \includegraphics[width=0.90\linewidth, trim={0 0.75cm 0 0},clip]{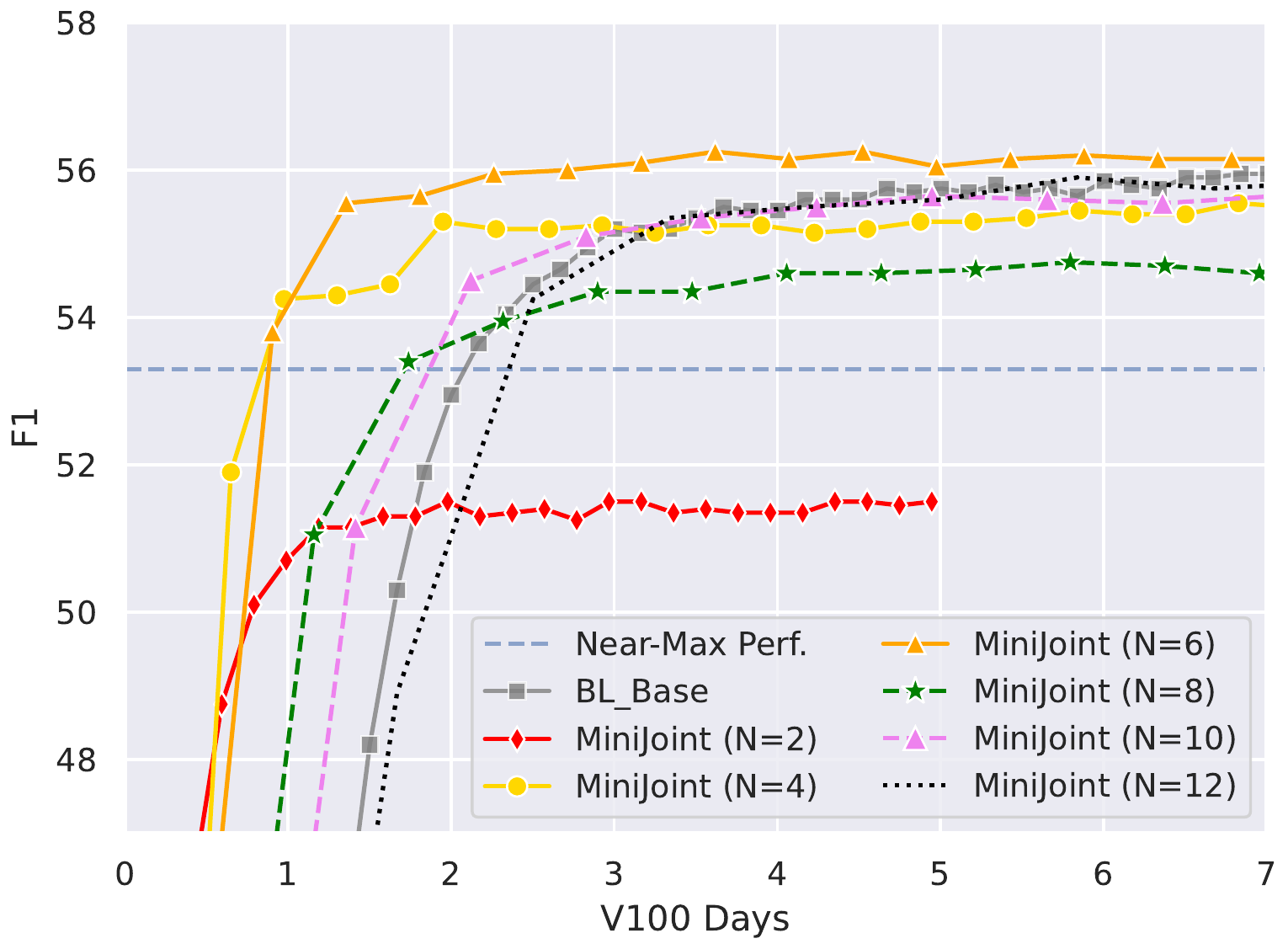}
    \caption{\textbf{MLQA.} (Arabic, German)}
    \label{fig:layerwise-mlqa-avg}
    \end{subfigure}
    
    \begin{subfigure}{\linewidth}
    \centering
    \includegraphics[width=0.90\linewidth, trim={0 0.75cm 0 0},clip]{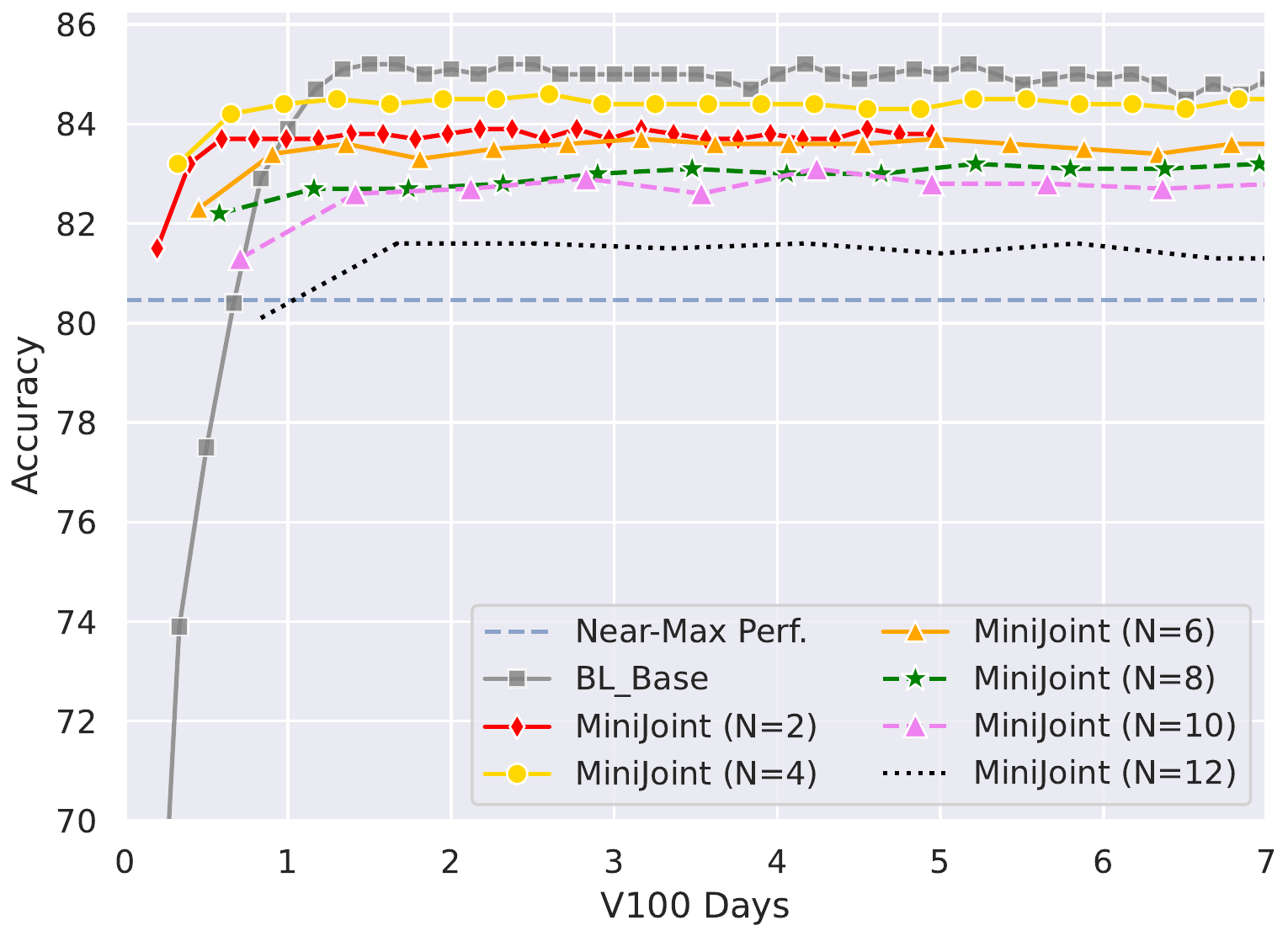}
    \caption{\textbf{PAWS-X.} (German only)}
    \label{fig:layerwise-paws-avg}
    \end{subfigure}
    
    \begin{subfigure}{\linewidth}
    \centering
    \includegraphics[width=0.90\linewidth, trim={0 0.1cm 0 0},clip]{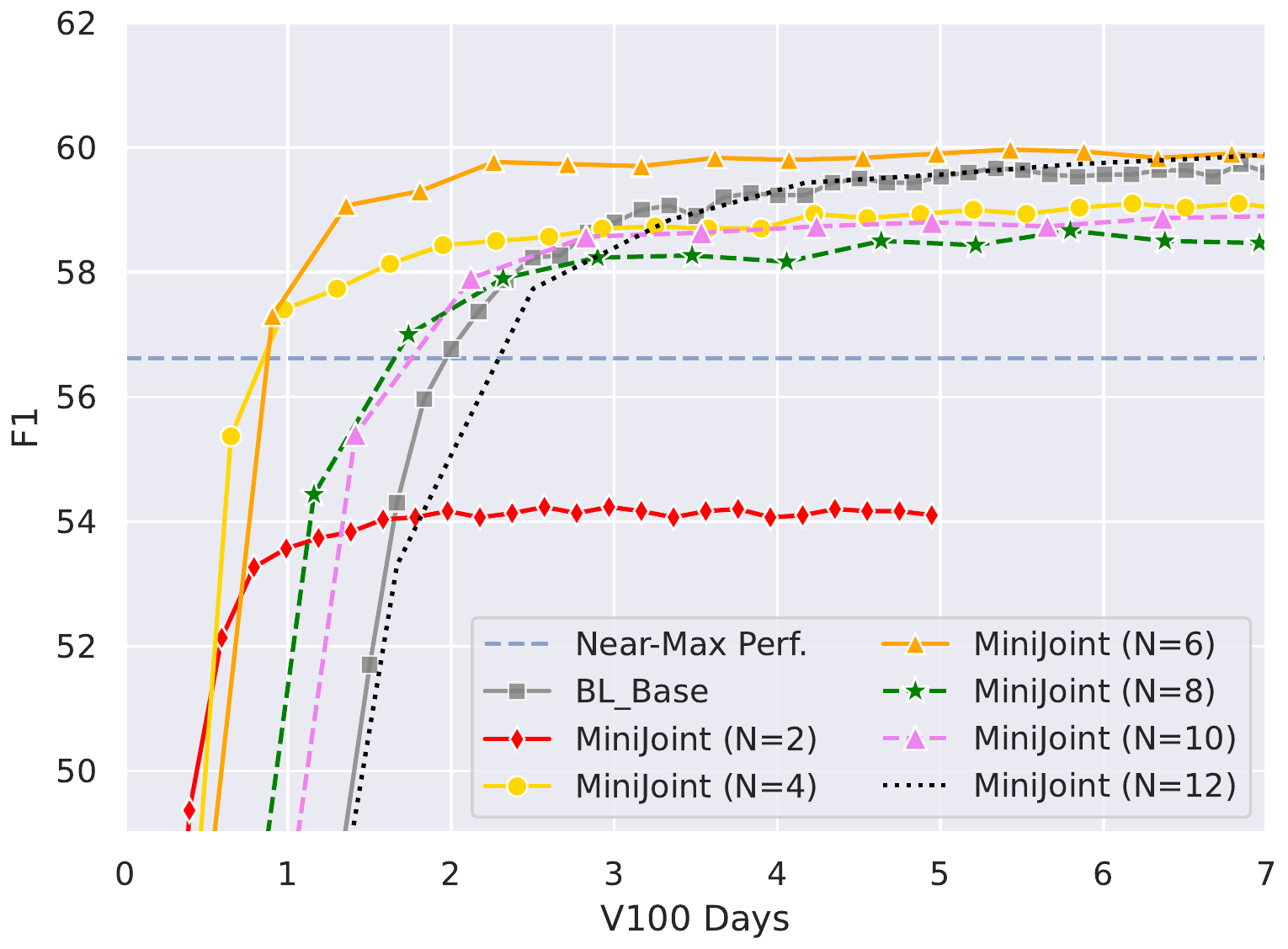}
    \caption{\textbf{XQuAD.} (Arabic, German, Turkish)}
    \label{fig:layerwise-xquad-avg}
    \end{subfigure}
    \caption{\textbf{Training curves for \adhoc{} with secondary head attached at varying layers.} Averaged over languages. Final performance in Table \ref{tab:layerwise-final}.}
    \label{fig:app-layerwise-avg}
\end{figure}

\begin{figure}
    \begin{subfigure}{\linewidth}
    \centering
    \includegraphics[width=0.92\linewidth, trim={0 0.75cm 0 0},clip]{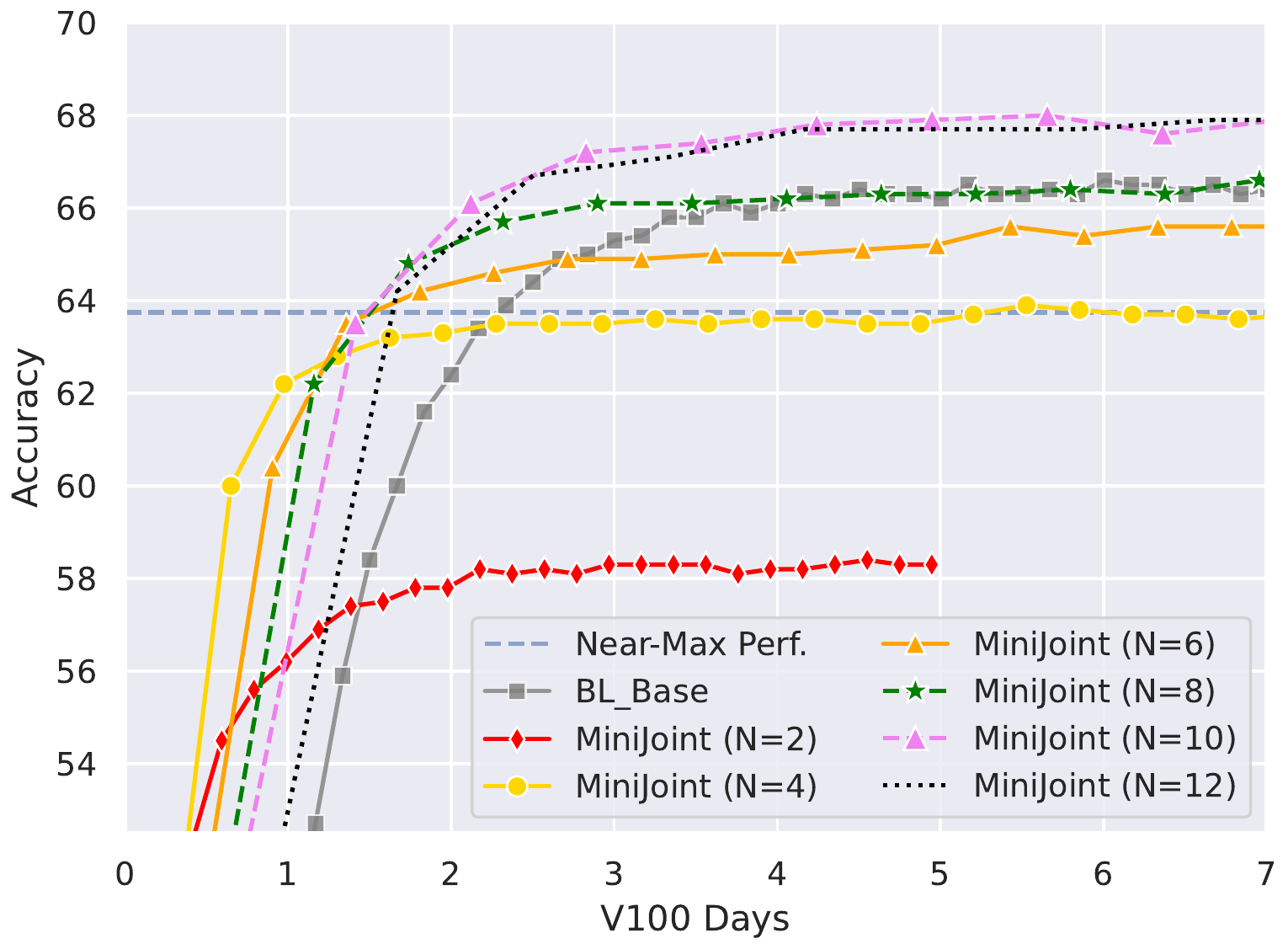}
    \caption{\textbf{XNLI.}}
    \label{fig:layerwise-xnli-tr}
    \end{subfigure}
    
     \begin{subfigure}{\linewidth}
    \centering
    \includegraphics[width=0.92\linewidth, trim={0 0.1cm 0 0},clip]{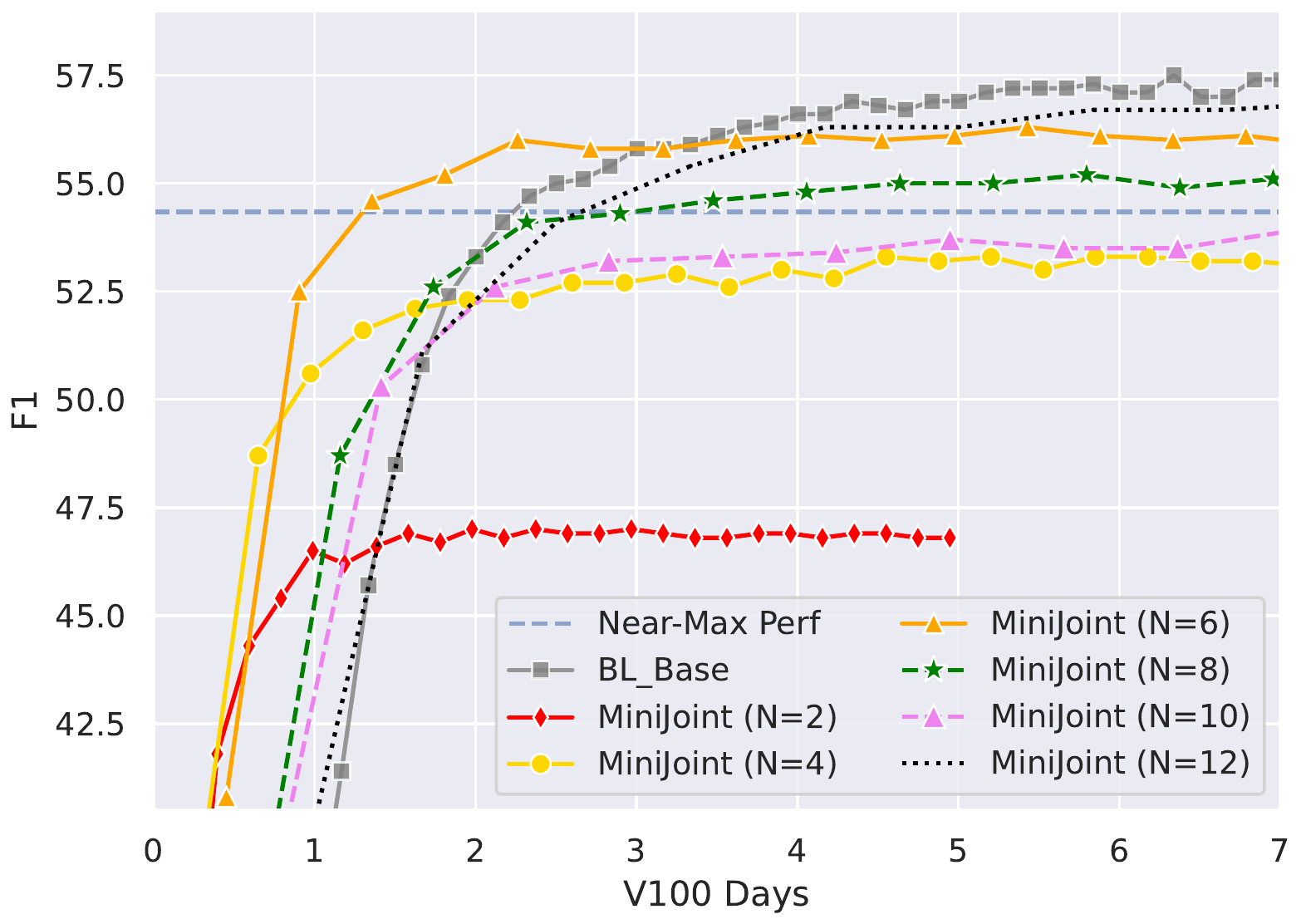}
    \caption{\textbf{XQuAD.}}
    \label{fig:layerwise-xquad-tr}
    \end{subfigure}
    \caption{\textbf{Training curves for \adhoc{} with secondary head attached at varying layers.} Turkish. Final performance in Table \ref{tab:layerwise-final}.}
    \label{fig:app-layerwise-tr}
\end{figure}

We extend the results of \S\ref{sec:layerwise} to MLQA, PAWS-X, and XQuAD, shown in Figure \ref{fig:app-layerwise-avg}. Figure \ref{fig:app-layerwise-tr} shows training curves for the particularly challenging language of Turkish on XNLI and XQuAD. Table \ref{tab:layerwise-final} shows performance at training completion.

\begin{table}[htb]
 \centering\footnotesize
 \setlength{\tabcolsep}{3pt}
\begin{tabular}{@{}cc|cccccc@{}}
\toprule
\multicolumn{1}{l}{} & \textbf{Figure:}  & \textbf{Fig. 4} & \textbf{A2(a)} & \textbf{A2(b)} & \textbf{A2(c)} & \textbf{A3(a)} & \textbf{A3(b)} \\ \midrule
\multirow{6}{*}{\textit{\textbf{N =}}}   & \textbf{2}        & 66.2            & 51.5           & 83.8           & 54.1           & 58.3           & 46.8           \\
                     & \textbf{4}        & 69.3            & 55.4           & 84.4           & 59.0           & 63.7           & 53.2           \\
                     & \textbf{6}        & 70.3            & 56.2           & 83.7           & 59.8           & 65.6           & 56.0           \\
                     & \textbf{8}        & 70.6            & 54.8           & 83.3           & 58.6           & 66.7           & 55.0           \\
                     & \textbf{10}       & 71.4            & 55.8           & 82.9           & 58.8           & 67.9           & 53.2           \\
                     & \textbf{12}       & 71.1            & 56.0           & 81.5           & 60.3           & 68.5           & 57.4           \\ \midrule
\multicolumn{1}{l}{} & \textbf{BL\_Base} & 71.2            & 56.1           & 84.7           & 59.6           & 67.1           & 57.2           \\ \bottomrule
\end{tabular}
\caption{\textbf{Performance at end of training for \adhoc{}.} Results correspond to Figures \ref{fig:layerwise-xnli-avg}, \ref{fig:app-layerwise-avg}, and \ref{fig:app-layerwise-tr}.}
\label{tab:layerwise-final}
\end{table}

\pagebreak
\subsection{Upper/Lower estimates on Time to Near-Maximal Performance}
\label{sec:app-near-max}
In \S\ref{sec:near-max}, we use linear interpolation to estimate GPU days to near-maximal performance if target performance occurred between checkpoints. In Table \ref{tab:app-layerwise-95}, we show the upper and lower estimates. Models are checkpointed every 5000 updates, so a lower estimate of 0.0 implies that the target score was achieved before first checkpoint. Because \adhoc{} with the secondary head attached at layer 4 was part of the main experiments, it was also checkpointed on steps 1000, 2000, 3000, and 4000. As such, estimates lower than 0.3 from this model imply that the target score was achieved hit before step 5000 (the first checkpoint for other models).

\begin{table*}
\centering\small
\begin{tabular}{lrlrrrrr}
\toprule
 &
  \multicolumn{1}{l}{\textbf{Layers:}} &
  \multicolumn{1}{c}{\textbf{2}} &
  \multicolumn{1}{c}{\textbf{4}} &
  \multicolumn{1}{c}{\textbf{6}} &
  \multicolumn{1}{c}{\textbf{8}} &
  \multicolumn{1}{c}{\textbf{10}} &
  \multicolumn{1}{c}{\textbf{12}} \\ \midrule
 &
  XNLI &
  \multicolumn{1}{r}{	
  \cellcolor[HTML]{65FF00}{0.4	(0.2	-	0.4)}} &
  \cellcolor[HTML]{93FF00}{0.5	(0.3	-	0.7)} &
  \cellcolor[HTML]{D9FF00}{0.7	(0.5	-	0.9)} &
  \cellcolor[HTML]{FEF705}{1.0	(0.6	-	1.2)} &
  \cellcolor[HTML]{FCEE0B}{1.1	(0.7	-	1.4)} &
  \cellcolor[HTML]{F8DB18}{1.4	(0.8	-	1.7)} \\
 &
  XQUAD &					
  \multicolumn{1}{r}{
  \cellcolor[HTML]{67FF00}{0.4	(0.2	-	0.4)}} &
  \cellcolor[HTML]{78FF00}{0.4	(0.3	-	0.7)} &
  \cellcolor[HTML]{BEFF00}{0.6	(0.5	-	0.9)} &
  \cellcolor[HTML]{FFFB03}{0.9	(0.6	-	1.2)} &
  \cellcolor[HTML]{E2FF00}{0.7	(0.7	-	1.4)} &
  \cellcolor[HTML]{FAE213}{1.3	(0.8	-	1.7)} \\
 &
  MLQA &				
  \multicolumn{1}{r}{
  \cellcolor[HTML]{D0FF00}{0.7	(0.6	-	0.8)}} &
  \cellcolor[HTML]{A7FF00}{0.6	(0.3	-	0.7)} &
  \cellcolor[HTML]{C1FF00}{0.6	(0.5	-	0.9)} &
  \cellcolor[HTML]{FDF308}{1.1	(0.6	-	1.2)} &
  \cellcolor[HTML]{FDF308}{1.0	(0.7	-	1.4)} &
  \cellcolor[HTML]{F8DB18}{1.4	(0.8	-	1.7)} \\
\multirow{-4}{*}{\textbf{de}} &
  PAWS &				
  \multicolumn{1}{r}{
  \cellcolor[HTML]{36FF00}{0.2	(0.0	-	0.2)}} &
  \cellcolor[HTML]{41FF00}{0.2	(0.2	-	0.3)} &
  \cellcolor[HTML]{83FF00}{0.4	(0.0	-	0.5)} &
  \cellcolor[HTML]{ABFF00}{0.5	(0.0	-	0.6)} &
  \cellcolor[HTML]{D6FF00}{0.7	(0.0	-	0.7)} &
  \cellcolor[HTML]{FDF308}{1.0	(0.8	-	1.7)} \\ \midrule
 &
  XNLI &
  \multicolumn{1}{c}{
  \cellcolor[HTML]{E34747}{$\infty$}} &  				
  \cellcolor[HTML]{FEFA03}{1.0	(0.7	-	1.0)} &
  \cellcolor[HTML]{FEFA03}{0.9	(0.9	-	1.4)} &
  \cellcolor[HTML]{FAE511}{1.3	(1.2	-	1.7)} &
  \cellcolor[HTML]{F3C427}{1.8	(1.4	-	2.1)} &
  \cellcolor[HTML]{EDA33D}{2.4	(1.7	-	2.5)} \\
 &
  XQUAD &
  \multicolumn{1}{c}{
  \cellcolor[HTML]{E34747}{$\infty$}} &			
  \cellcolor[HTML]{EBFF00}{0.8	(0.7	-	1.0)} &
  \cellcolor[HTML]{FFFE01}{0.9	(0.5	-	0.9)} &
  \cellcolor[HTML]{F6D21E}{1.6	(1.2	-	1.7)} &
  \cellcolor[HTML]{F4C527}{1.8	(1.4	-	2.1)} &
  \cellcolor[HTML]{ECA13E}{2.4	(1.7	-	2.5)} \\
\multirow{-3}{*}{\textbf{ar}} &
  MLQA &
  \multicolumn{1}{c}{
  \cellcolor[HTML]{E34747}{$\infty$}} &				
  \cellcolor[HTML]{FEF705}{1.0	(1.0	-	1.3)} &
  \cellcolor[HTML]{FDF308}{1.1	(0.9	-	1.4)} &
  \cellcolor[HTML]{EFAE36}{2.2	(1.7	-	2.3)} &
  \cellcolor[HTML]{F1B730}{2.1	(1.4	-	2.1)} &
  \cellcolor[HTML]{EC9D41}{2.5	(1.7	-	2.5)} \\ \midrule
 &
  XNLI &
  \multicolumn{1}{c}{
  \cellcolor[HTML]{E34747}{$\infty$}} &			
  \cellcolor[HTML]{E06666}{5.3	(5.2	-	5.5)} &
  \cellcolor[HTML]{F8D919}{1.5	(1.4	-	1.8)} &
  \cellcolor[HTML]{F8D919}{1.5	(1.2	-	1.7)} &
  \cellcolor[HTML]{F8D919}{1.5	(1.4	-	2.1)} &
  \cellcolor[HTML]{F6CF20}{1.6	(0.8	-	1.7)} \\
\multirow{-2}{*}{\textbf{tr}} &
  XQUAD &
  \multicolumn{1}{c}{\cellcolor[HTML]{E34747}{$\infty$}} &
  \multicolumn{1}{c}{\cellcolor[HTML]{E34747}{$\infty$}} &
  \cellcolor[HTML]{FAE511}{1.3	(0.9	-	1.4)} &
  \cellcolor[HTML]{E68352}{3.0	(2.9	-	3.5)} &
  \multicolumn{1}{c}{\cellcolor[HTML]{E34747}{$\infty$}} &
  \cellcolor[HTML]{EA9447}{2.7	(2.5	-	3.3)} \\
  \bottomrule
\end{tabular}
\caption{\textbf{Estimated V100 days to near-maximal performance (see \S\ref{sec:near-max}) for \adhoc{} with secondary head attached at varying layers.} $\downarrow$ better. $\infty$: never hit target performance. (lower - upper) bounds on the estimate.}
\label{tab:app-layerwise-95}
\end{table*}

\subsection{Variance across Pretraining Runs} \label{app:pretraining-seed}
While we average results over 5 finetuning runs, we always use the same pretrained model. Early in development, we noticed that there could be a difference between different pretraining runs. While it was not feasible to repeat all experiments with different pretraining seeds due to computational cost, we performed 3 additional runs of \baselinelarge{} for Arabic. We see a difference up to 3 points across runs in Table \ref{tab:app-seeding}. This is task dependent, as the best run on XNLI is the worst on MLQA.
\begin{table}[htb]
\centering
\begin{tabular}{ccc}
\toprule
  & XNLI & MLQA \\
  \midrule
Run \#1         & 67.7 & 51.4 \\
Run \#2             & 69.3 & 51.1 \\
Run \#3             & 68.7 & \textbf{52.7} \\
\midrule
Main run             & \textbf{69.6} &  49.6 \\
\bottomrule
\end{tabular}
\caption{\textbf{Arabic development performance for \baselinelarge{} with different pretraining seeds.} Results averaged over 5 finetuning runs.}
\label{tab:app-seeding}
\end{table}

\end{document}